\documentclass[]{bytedance}
\usepackage[toc,page,header]{appendix}

\usepackage{graphicx}
\usepackage{listings}
\usepackage{booktabs}
\usepackage{hyperref}
\hypersetup{colorlinks=true,breaklinks,linkcolor=red,citecolor=cyan}
\usepackage{url}
\usepackage{float}
\usepackage{amsfonts}
\usepackage{adjustbox}
\usepackage{wrapfig}
\usepackage[table,dvipsnames]{xcolor}

\title{
\adjustbox{valign=c}{\includegraphics[height=1.5em]{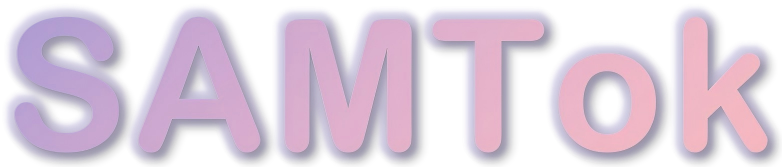}}\\
\vspace{3mm}
  SAMTok: Representing Any Mask with Two Words
}

{\small
\author[\star]{Yikang Zhou$^{1,2~}$}
\author[\star]{Tao Zhang$^{1,2~}$}
\author[]{Dengxian Gong$^{1~}$}
\author[]{Yuanzheng Wu$^{1~}$}
\author[]{Ye Tian$^{2~}$}
\author[]{Haochen Wang$^{2~}$}
\author[]{Haobo Yuan$^{2~}$}
\author[]{Jiacong Wang$^{2~}$}
\author[]{Lu Qi$^{1~}$}
\author[]{Hao Fei$^{3~}$}
\author[]{Anran Wang$^{2~}$}
\author[]{Zhuochen Wang$^{2~}$}
\author[]{Yujing Wang$^{2~}$}
\author[]{Cheng Chen$^{2~}$}
\author[\dagger]{\\Shunping Ji$^{1~}$}
\author[\ddagger]{Xiangtai Li$^{2~}$}
}
\affiliation[]{ 
{Wuhan University$^{1}$} \quad  ByteDance$^{2}$ \quad  NUS$^{3}$ \\
\vspace{3pt}
{$^\star$: Equal contributions~ $^\dagger$: Corresponding Author~ $^\ddagger$: Project Leader} \\
\vspace{3pt}
{Emails: \email{xiangtai.li@bytedance.com}, \email{zhouyik@whu.edu.cn}, \email{zhang\_tao@whu.edu.cn}} \\
\vspace{3pt}
{Project Page: \ \url{https://zhouyiks.github.io/projects/SAMTok/}}
}

\abstract{
Pixel-wise capabilities are essential for building interactive intelligent systems.
However, pixel-wise multi-modal LLMs (MLLMs) remain difficult to scale due to complex region-level encoders, specialized segmentation decoders, and incompatible training objectives.
To address these challenges, we present \textbf{SAMTok}, a discrete mask tokenizer that converts any region mask into two special tokens and reconstructs the mask using these tokens with high fidelity.
By treating masks as new language tokens, \textbf{SAMTok} enables base MLLMs (such as the QwenVL series) to learn pixel-wise capabilities through standard next-token prediction and simple reinforcement learning, without architectural modifications and specialized loss design.
\textbf{SAMTok} builds on SAM2 and is trained on 209M diverse masks using a mask encoder and residual vector quantizer to produce discrete, compact, and information-rich tokens.
With 5M SAMTok-formatted mask understanding and generation data samples, \textbf{QwenVL-SAMTok} attains state-of-the-art or comparable results on region captioning, region VQA, grounded conversation, referring segmentation, scene graph parsing, and multi-round interactive segmentation.
We further introduce a textual answer-matching reward that enables efficient reinforcement learning for mask generation, delivering substantial improvements on GRES and GCG benchmarks. 
Our results demonstrate a scalable and straightforward paradigm for equipping MLLMs with strong pixel-wise capabilities.
Our code and models are available.
}

\date{\today}

\begin{document}
\maketitle
\begingroup  
\renewcommand{\thefootnote}{}
\footnotetext{This work was done by Yikang Zhou and Tao Zhang during their internships at ByteDance.}  
\endgroup

\section{Introduction}
\label{sec:intro}
%
%
%

\begin{figure}[t]
\centering
\includegraphics[width=0.99\textwidth]{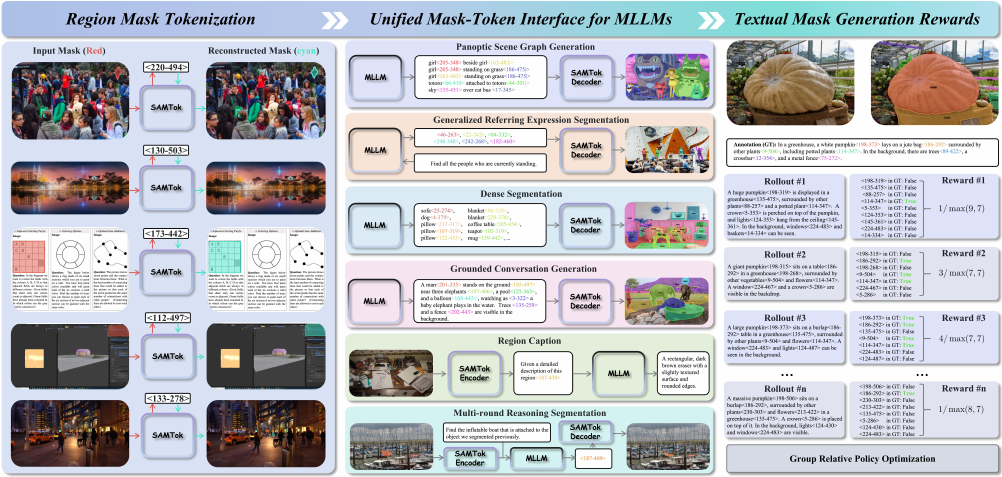}
\captionof{figure}{\textbf{SAMTok} provides a simple yet unified mask-token interface for MLLMs. \textbf{(Left)} SAMTok compresses region masks into two discrete tokens and faithfully reconstructs them across diverse visual domains. \textbf{(Middle)} Injecting these mask tokens into MLLMs enables a wide range of region-level mask generation and understanding tasks. \textbf{(Right)} The text-based representation of region masks allows a purely textual answer-matching reward for the GRPO of the mask generation task.}
\label{fig:teaser}
\end{figure}

Multi-modal large language models~\cite{qwenvl, qwen2vl, qwen25vl, internvl, internvl25, internvl3, internvl35, deepseekvl2, glm45v, gpt4o, claude37, gemini} have made significant progress in real-world applications, driven by rapid advancements in language models.
Pixel-wise large language models~\cite{sa2va, gar, dam, pixelrefer,himtok, bigverdi2025perception} have also drawn considerable research efforts, with most focusing on solving fine-grained visual language tasks, which are essential for building interactive intelligent systems.

However, currently, building a good pixel-wise MLLM with strong scalability faces four challenges:
(1) There are no unified designs to handle both mask-in and mask-out inputs.
Existing models rely on complex region-level feature pooling designs~\cite{osprey, sa2va, dam, gar}, while mask output depends on carefully designed segmentation decoders~\cite{lisa, zhang2024omg, sa2va, hyperseg, padt}.
Although unified modeling can be achieved through alternative approaches such as bounding boxes or points~\cite{segzero, visionreasoner}, this comes at the cost of reduced precision and introduced ambiguity.
(2) Current state-of-the-art pixel-wise MLLMs~\cite{sa2va, hyperseg, lens} \textit{cannot} directly and concisely apply reinforcement learning (RL) to mask generation tasks since they use continuous embeddings to connect the MLLM with the segmentation head. 
(3) Specially designed modules added for mask understanding and generation capabilities typically require co-training with the MLLM~\cite{sa2va, zhang2024omg, padt, hyperseg, lens, himtok}. 
In addition, the different training losses and forward pipelines introduce substantial complexity for scaling training with VQA and pure text data.
(4) Meanwhile, several explorations~\cite{argenseg, lan2024text4seg, alto, wang2023visionllm} attempt to circumvent these issues by treating masks as special image tokens or representing them as text in formats similar to RLE encoding or polygons.
However, these typically incur enormous inference costs, with a single mask being represented by dozens or even hundreds of tokens.
Therefore, we pose one essential question: ``\textbf{\textit{How can we non-intrusively endow base MLLMs (such as the QwenVL series) with pixel-wise capabilities, making the learning process as simple as VQA training—requiring only next-token prediction loss for supervised fine-tuning (SFT) and straightforward reinforcement learning (RL)?}}"

%


In this paper, we propose SAMTok, a discrete mask tokenizer that tokenizes masks into textual special words (text tokens) and detokenizes these textual special words into masks, thereby transforming masks into a new language for MLLMs to learn from, similar to regular text data.
As shown in Fig.~\ref{fig:teaser}, our proposed SAMTok can convert diverse masks into textual special tokens and accurately reconstruct the corresponding masks.
Through SAMTok, any MLLM can acquire powerful pixel-wise capabilities by learning like language data through supervised fine-tuning and reinforcement learning, without any additional architectural modifications or specialized loss design.

SAMTok is initialized from the foundation segmentation model SAM2~\cite{sam2} to accelerate convergence.
We incorporate several additional components, including a mask encoder and a residual vector quantizer, to encode masks into compact, information-rich mask embeddings and to discretize the mask embeddings into two discrete tokens, respectively.
%
To obtain strong mask generation capabilities, we train SAMTok on over 209M masks, drawn from diverse existing segmentation datasets.
As a result, SAMTok can tokenize any region masks and reconstruct them better as shown on the left of Fig.~\ref{fig:teaser}.

To further validate SAMTok for pixel-understanding tasks, we collect approximately 5M diverse samples of mask-understanding and generation data.
These data are converted into a standard vision question answering format using SAMTok, which tokenizes masks as special text, thereby transforming them into a conventional format containing only images and text.
We use these data to perform supervised fine-tuning on the QwenVL series~\cite{qwen2vl, qwen25vl} models.
We are surprised to find that MLLMs can learn pixel-wise capabilities just like learning language, without any fancy design, as shown in the middle of Fig.~\ref{fig:teaser}.

QwenVL-SAMTok achieves stronger performance or performance comparable to carefully designed expert MLLMs on mask understanding tasks, including region captioning and region VQA~\cite{mdpv, videorefer, osprey, dam}, mask generation tasks, including grounded conversation generation~\cite{glamm}, referring segmentation~\cite{groundingsuite, gres, refcoco}, and scene graph parsing~\cite{psg}, as well as interleaved mask understanding and generation tasks such as multi-round interactive segmentation~\cite{segllm}.

Thanks to SAMTok's efficient representation of masks as discrete textual special tokens, we can explore the upper bound of MLLM performance in mask generation using a simple yet effective RL approach.
Specifically, as illustrated in the right of Fig.~\ref{fig:teaser}, we propose a textual answer-matching reward function for generated masks, which is better aligned with reward mechanisms in the language domain.
After applying GRPO~\cite{grpo}, QwenVL-SAMTok demonstrates significant improvements on the GRES~\cite{gres} validation set, achieving gains of 8.9\% in gIoU and 21.0\% in N-acc, as well as on the GCG~\cite{glamm} validation set with improvements of 4.7\% in AP50 and 6.6\% in Recall. 
These results surpass the previous SOTA methods by 4.3\% in gIoU, 8.3\% in N-acc, 8.3\% in AP50, and 8.4\% in Recall.

In summary, our contributions are as follows:

\begin{itemize}
\item We propose a novel paradigm for MLLMs to model masks as a new language, enabling them to learn mask understanding and generation capabilities just like natural language without requiring architecture modifications or additional loss design.

\item We propose SAMTok, which can accurately achieve bidirectional conversion between masks and textual special tokens. Based on SAMTok, the QwenVL series of MLLMs acquire strong pixel-wise capabilities through next token prediction loss, achieving SOTA performance across dozens of diverse benchmarks.

\item We design a textual answer-matching reward function that enables MLLMs to perform reinforcement learning on mask generation tasks similar to natural language data, demonstrating significant performance improvements.

\end{itemize}
\begin{figure}[t]
\centering
\includegraphics[width=0.99\textwidth]{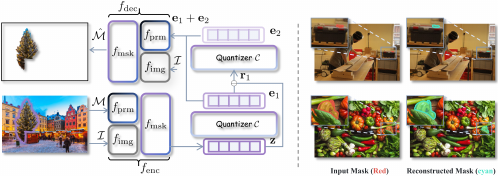}
\caption{ Our SAMTok architecture \textbf{(Left)} and mask reconstruction examples \textbf{(Right)}. SAMTok has an encoder $f_{\text{enc}}$, a vector quantizer with codebook $\mathcal{C}$, and a decoder $f_{\text{dec}}$. Both $f_{\text{enc}}$ and $f_{\text{dec}}$ are instantiated with a SAM model, which includes an image backbone $f_{\text{img}}$, a prompt encoder $f_{\text{prm}}$, and a mask decoder $f_{\text{msk}}$. 
Given an input image $\mathcal{I}$ and a region $\mathcal{M}$ (e.g., the area outlined in purple), the SAMTok encoder $f_{\text{enc}}$ first encodes the 2D mask into a mask embedding $\mathbf{z}$, then performs two-stage quantization to obtain a discrete mask embedding $[\mathbf{e}_1, \mathbf{e}_2]$. 
The SAMTok decoder $f_{\text{dec}}$ reconstructs the 2D mask $\hat{\mathcal{M}}$ from the original image and the region’s discrete mask embeddings.}
\label{fig:samtok}
\end{figure}

\section{Method}
\label{sec:method}

\subsection{SAMTok}
\label{sec:method-samtok}
An ideal region tokenizer should possess the following characteristics to better fit into LLM processing:
(1) the ability to convert between 2D masks and latent region representations, enabling MLLMs to both input and output regions;
(2) efficient and accurate region representations that minimize inference cost while maintaining precision; and
(3) discrete region representations to facilitate reinforcement learning.

Existing methods have demonstrated complementary strengths in different aspects:
(1) Variational autoencoders (VAE) excel at the property of converting between images and latent representations.
(2) Perception models like SAM excel at the property of accurately segmenting objects through a single embedding.
(3) Vector quantization methods excel at the property of effectively discretizing continuous latents into compact codes to facilitate RL training.
SAMTok is a mask VAE integrated with vector quantization, where the mask latent features are represented as a single, highly condensed embedding mirroring the embedding characteristics of perception models.
As shown in Fig.~\ref{fig:samtok}, SAMTok has an encoder $f_{\text{enc}}$, a vector quantizer with codebook $\mathcal{C}$, and a decoder $f_{\text{dec}}$.
Both SAMTok encoder $f_{\text{enc}}$ and decoder $f_{\text{dec}}$ are instantiated with a SAM model, which includes an image backbone $f_{\text{img}}$, a prompt encoder $f_{\text{prm}}$, and a mask decoder $f_{\text{msk}}$. 
The details of each component of SAMTok will be elaborated below.

\noindent\textbf{SAMTok Encoder.} Our encoder $f_{\text{enc}}$ is a SAM model~\cite{sam} with the final mask prediction head removed from the SAM mask decoder $f_{\text{msk}}$. 
Analogous to interactive segmentation, the SAM prompt encoder $f_{\text{prm}}$ encodes a 2D mask $\mathcal{M}$ into dense prompt embeddings that share the same spatial resolution as the image features encoded by SAM’s image backbone $f_{\text{img}}$ for the current image $\mathcal{I}$.
We add these dense prompt embeddings to the image features, then feed the result into the SAM mask decoder $f_{\text{msk}}$, where it interacts with a pre-initialized mask embedding. 
At this point, we obtain the desired SAMTok encoder output: a ${d}$-dimensional continuous mask embedding ${\mathbf{z}}$:
\begin{equation}
\mathbf{z} = f_{\text{enc}}(\mathcal{I}, \mathcal{M}) = f_{\text{msk}}(f_{\text{img}}(\mathcal{I}), f_{\text{prm}}(\mathcal{M})) \in \mathbb{R}^{d}
\label{eq:samtok-enc}
\end{equation}

\noindent\textbf{Quantizer.}
We employ a residual quantization scheme~\cite{rq} to discretize the continuous mask embedding $\mathbf{z}$, since it can achieve high-fidelity quantization with a compact codebook $\mathcal{C}$ relative to other quantization schemes~\cite{vqvae,rq,fsq}.
The continuous mask embedding $\mathbf{z}$ is processed through two quantization steps, as shown in Eq.~\ref{eq:samtok-quant}.
First, we conduct a nearest-neighbor lookup in the codebook $\mathcal{C}$ for the continuous mask embedding $\mathbf{z}$ and compute the residual embedding $\mathbf{r}_1$ between the continuous mask embedding $\mathbf{z}$ and the retrieved latent embedding $\mathbf{e}_1$.
Second, we perform a nearest-neighbor lookup in the codebook $\mathcal{C}$ for the residual embedding $\mathbf{r}_1$.
By applying residual quantization, we achieve high-fidelity compression of the region mask, which can then be reconstructed by the SAMTok decoder.
\begin{equation}
\left\{
\begin{array}{r@{\;}l}
\mathbf{e}_1 &= \operatorname*{argmin}_{\mathbf{e} \in \mathcal{C}} \|\mathbf{z} - \mathbf{e}\|_2^2, \\[0.8ex]
\mathbf{r}_1 &= \mathbf{z} - \mathbf{e}_1, \\[0.8ex]
\mathbf{e}_2 &= \operatorname*{argmin}_{\mathbf{e} \in \mathcal{C}} \|\mathbf{r}_1 - \mathbf{e}\|_2^2, \\[0.8ex]
\mathbf{q}~ &= [\mathbf{e}_1, \mathbf{e}_2]
\end{array}
\right.
\label{eq:samtok-quant}
\end{equation}

\noindent\textbf{SAMTok decoder.}
Our decoder $f_{\text{dec}}$ is a full SAM model~\cite{sam}.
We treat the discrete mask embeddings $[\mathbf{e}_1, \mathbf{e}_2]$ as special language prompt embeddings~\cite{lisa,sa2va} for the current image $\mathcal{I}$, and require SAM to segment the corresponding region mask (Eq.~\ref{eq:samtok-dec}). 
Inside the SAM prompt encoder $f_{\text{prm}}$, we directly sum all discrete mask embeddings to form a sparse prompt embedding. 
In the SAM mask decoder $f_{\text{msk}}$, the pre-initialized mask embedding interacts with the sparse prompt embedding via self-attention and with the image features via cross-attention, ultimately recovering the features of the continuous mask embedding $\mathbf{z}$. 
Applying the mask prediction head then reconstructs the 2D region mask $\hat{\mathcal{M}}$.
\begin{equation}
\hat{\mathcal{M}} = f_{\text{dec}}(\mathcal{I}, [\mathbf{e}_1, \mathbf{e}_2]) = f_{\text{msk}}(f_{\text{img}}(\mathcal{I}), f_{\text{prm}}([\mathbf{e}_1, \mathbf{e}_2]))
\label{eq:samtok-dec}
\end{equation}

\noindent\textbf{Training of SAMTok.}
SAMTok is trained for the mask reconstruction task.
We collect large-scale mask data from open-source datasets~\cite{sam}, totaling 209 million masks for training.
The training objective follows~\cite{rq}, including a reconstruction loss and a commitment loss.
The reconstruction loss is computed as the sum of the cross-entropy loss and the Dice loss between the reconstructed and original masks.
\begin{equation}
\left\{
\begin{aligned}
\mathcal{L} &= \mathcal{L}_{\text{recon}} + \lambda \mathcal{L}_{\text{commit}}, \\
\mathcal{L}_{\text{recon}} &= \mathcal{L}_{\text{CE}}(\mathcal{M}, \hat{\mathcal{M}}) + \mathcal{L}_{\text{DICE}}(\mathcal{M}, \hat{\mathcal{M}}), \\
\mathcal{L}_{\text{commit}} &= \|\mathbf{z} - \operatorname{sg}(\mathbf{e}_1)\|_2^2 + \|\mathbf{r}_1 - \operatorname{sg}(\mathbf{e}_2)\|_2^2,
\end{aligned}
\right.
\label{eq:samtok-loss}
\end{equation}
where $\operatorname{sg}(\cdot)$ denotes the stop-gradient operator, and $\lambda$ controls the quantization commitment.

\subsection{Unified Mask-Token Interface for MLLMs}
\label{sec:method-vlm}

\begin{figure*}[t]
\centering
\includegraphics[width=0.99\textwidth]{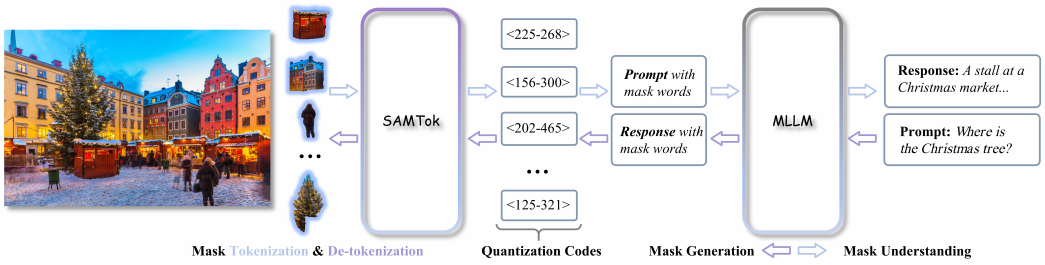}
\caption{ Unified mask-token interface for MLLMs. For the mask understanding task (left to right), SAMTok first tokenizes region masks into quantization codes, then formats them into mask words, which are used in the MLLM prompt to refer to the corresponding image regions. 
For the mask generation task (right-to-left), the MLLM first produces mask words according to the instruction, then maps them to quantization codes, after which SAMTok reconstructs the 2D masks.}
\label{fig:mllm}
\end{figure*}

Through SAMTok, any region mask can be converted into two discrete tokens.
This enables MLLMs to understand and predict masks in a text-based manner.
Specifically, we treat mask tokens as a new language and add mask special tokens to MLLM's vocabulary, equal in number to the SAMTok codebook size.
Any region can then be represented as a pair of mask special tokens for the MLLM to understand and generate.
This paradigm enables MLLMs to learn region-mask understanding and generation capabilities through simple next-token prediction, just as they do with text.
This training process no longer relies on any task-specific losses, such as the segmentation losses used in classical region-level MLLMs~\cite{lisa, sa2va, zhang2024omg, himtok}, nor does it require any architectural modifications to the MLLM.

\noindent\textbf{Mask Understanding.}
SAMTok can tokenize masks of arbitrary image regions into compact, information-dense mask tokens.
These mask tokens serve as efficient and accurate references to region masks for MLLM input.
For example, in region captioning tasks, the input contains images, text instructions, and masks. 
The SAMTok encoder first encodes masks into quantization codes, which are then formatted into predefined special tokens.
Then, we insert these special tokens into the text instruction, analogous to using a text box but more efficient and easier for MLLMs to understand.

\noindent\textbf{Mask Generation.}
SAMTok can convert discrete mask tokens back into explicit segmentation masks.
Consequently, MLLMs can achieve mask generation capabilities by predicting predefined special tokens, such as for referring segmentation tasks.
When visualization of MLLM-generated masks is required during inference, the SAMTok decodes the mask special tokens in the response into segmentation masks.
Specifically, the predicted special tokens are first converted into quantization codes, which are then used to retrieve the corresponding latent embeddings from the codebook.
These latent embeddings are subsequently fed into the SAMTok decoder to decode the segmentation mask.

\noindent\textbf{Co-training of the MLLM.}
%
%
After SAMTok tokenizes all region masks into words, all mask-related tasks can be preprocessed into purely textual corpora, including mask-to-text, text-to-mask, interleaved text–mask generation, and interactive tasks (where masks serve as both inputs and outputs). 
In particular, we unify a diverse set of tasks—such as grounded conversation generation (GCG), panoptic scene graph generation (PSG), region captioning, region-level VQA, generalized referring expression segmentation (GRES), and visual grounding—under a single textual formulation.
For example, to convert region captioning data into a purely textual corpus, region masks are first transformed into mask words and inserted into the textual prompt, which is then paired with the corresponding region caption to form a single dialog turn. Similarly, to preprocess GCG data, all region masks are converted into mask words and interleaved into the image caption immediately after the corresponding phrases, together with an instruction prompt, to construct a unified conversational sequence.
Consequently, under any multimodal training framework, all these tasks can be co-trained using the standard next-token prediction loss, without introducing any customized loss functions or architectural modifications.

\subsection{Reinforcement Learning for Mask Generation}
\label{sec:method-rl}

The design of SAMTok is naturally compatible with RL, and we leverage RL to explore its upper bound.
However, enhancing MLLM's mask generation capabilities through RL poses significant challenges~\cite{visionreasoner, lens, segr1}.
In particular, many methods~\cite{lisa,sa2va,hyperseg,yuan2025visual} that pass segmentation embeddings can achieve strong performance via SFT, they struggle to benefit from further RL-based improvements due to their reliance on continuous features.
Although several works~\cite{visionreasoner,lens,segzero} have employed RL by predicting bounding boxes or points as auxiliary signals, they depend on models like SAM~\cite{sam} to convert these boxes or points into masks for reward computation, which substantially increases the overall system complexity.

SAMTok successfully represents masks using special words, providing substantial benefits for RL:
(1) The discrete textual representation of masks enables direct application of established RL algorithms.
(2) Mask reward evaluation can be performed through simple character matching without requiring additional tools to extract explicit masks.
In this work, we adopt group relative policy optimization (GRPO)~\cite{grpo} as the RL algorithm to optimize mask generation, as GRPO has been widely adopted by recent pixel-level MLLMs~\cite{lens,visionreasoner, segzero, segr1, alto} in multimodal reinforcement learning settings.

Specifically, during training, we first extract all special mask words from the model's rollout response (as shown in the right part of Fig.~\ref{fig:teaser}).
After removing duplicates, we check whether each mask word appears in the ground truth answer string. If it does, we count it as a true positive (TP).
We then use the ratio of TPs to the total number of mask words as the reward:
\begin{equation}
\mathcal{R}_\text{mask} = \mathcal{N}_\text{TP} / \operatorname*{max}(\mathcal{N}_{pred}, \mathcal{N}_{gt}),
\label{eq:rl-reward}
\end{equation}
where $\mathcal{N}_\text{TP}$ is the number of predicted true positive masks, $\mathcal{N}_{pred}$ is the number of predicted masks without deduplication to penalize repetitive predictions, and $\mathcal{N}_{gt}$ is the number of ground truth masks.

\begin{figure*}[t]
\centering
\includegraphics[width=0.99\textwidth]{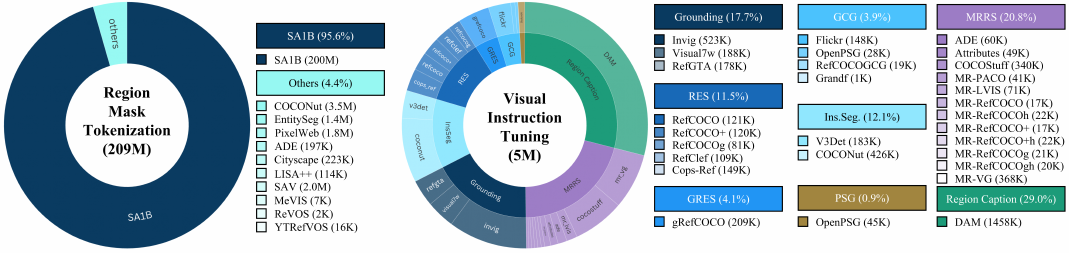}
\caption{Overview of dataset used to train the SAMTok (left) and MLLM (right). We use 209M masks to train SAMTok, and 5M conversations to fine tune MLLMs.}
\label{fig:dataset}
\end{figure*}

\begin{table*}[t!]
    \centering
    \caption{Results on interleaved text-mask generation task (GCG). ``(ft)''  indicates models further finetuned on GCG after mixed training. The best results are \textbf{bold} and the second-best results are \underline{underlined} (excluding our RL results). ``(rl)''  indicates models further trained with GRPO after mixed training.}
    \label{tab:exp-gcg}
    \setlength{\tabcolsep}{5pt}
    \resizebox{0.95\textwidth}{!}{
    \begin{tabular}{l | c | c c c c c | c c c c c}
    \toprule[1.5pt]
        \multirow{2}{*}{\textbf{Method}} & \multirow{2}{*}{\textbf{Size}} & \multicolumn{5}{c|}{\textbf{Val}} & \multicolumn{5}{c}{\textbf{Test}} \\
        ~ & ~ & \textbf{METEOR} & \textbf{CIDEr} & \textbf{AP50} & \textbf{mIoU} & \textbf{Recall} & \textbf{METEOR} & \textbf{CIDEr} & \textbf{AP50} & \textbf{mIoU} & \textbf{Recall} \\
        \hline
LISA~\cite{lisa} & 7B & 13.0 & 33.9 & 25.2 & 62.0 & 36.3 & 12.9 & 32.2 & 24.8 & 61.7 & 35.5\\
GLaMM~\cite{glamm} & 7B & 16.2 & 47.2 & 30.8 & 66.3 & 41.8 & 15.8 & 43.5 & 29.2 & 65.6 & 40.8\\
OMG-LLaVA~\cite{zhang2024omg} & 7B & 14.9 & 41.2 & 29.9 & 65.6 & -- & 14.5 & 38.5 & 28.6 & 64.7 & --\\
Sa2VA~\cite{sa2va} & 8B & 16.4 & 49.5 & 33.2 & 67.7 & 45.1 & 16.2 & 49.0 & 32.2 & 66.8 & 44.5\\
\rowcolor{blue!5}\textbf{Qwen25VL-SAMTok} & 3B & 16.9 & 51.8 & 36.8 & 71.7 & 46.9 & 16.7 & 52.5 & 36.1 & 70.9 & 47.0\\
\rowcolor{blue!5}\textbf{Qwen25VL-SAMTok (ft)} & 3B & \underline{17.3} & 54.7 & 37.0 & 71.7 & 47.7 & \underline{17.0} & \underline{53.9} & 36.3 & \underline{71.4} & 48.0 \\
\rowcolor{blue!5}\textbf{Qwen3VL-SAMTok} & 4B & 16.1 & 49.5 & 37.8 & 72.1 & 47.6 & 16.1 & 52.0 & \underline{37.4} & 70.6 & \underline{48.2} \\
\rowcolor{blue!5}\textbf{Qwen25VL-SAMTok} & 7B & 17.2 & \underline{54.8} & \underline{38.2} & \underline{72.6} & \underline{48.9} & \underline{17.0} & \textbf{54.5} & 37.0 & \textbf{71.7} & 47.9 \\
\rowcolor{blue!5}\textbf{Qwen25VL-SAMTok (ft)} & 7B & \textbf{17.7} & \textbf{55.0} & \textbf{38.5} & \textbf{72.9} & \textbf{49.8} & \textbf{17.4} & 53.7 & \textbf{37.5} & 71.2 & \textbf{48.7} \\
        \hline
\rowcolor{blue!5}\textbf{Qwen25VL-SAMTok (rl)} & 3B & 16.9 & 52.5 & 41.5 & 73.5 & 53.5 & 16.5 & 51.0 & 40.4 & 73.2 & 53.5 \\
    \bottomrule[1.5pt]
    \end{tabular}
    }
\end{table*}

\begin{table*}[t!]
    \centering
    \caption{Results on multi-round interactive segmentation tasks: including MR-RefCOCO/+/g (object-level) and MR-PACO (part-level). The evaluation metric is cIoU. We report average cIoU across MR-RefCOCO/+/g.}
    \label{tab:exp-segllm}
    \setlength{\tabcolsep}{3pt}
    \resizebox{0.95\textwidth}{!}{
    \begin{tabular}{l | c | c c c c c | c c c c}
    \toprule[1.5pt]
        \multirow{2}{*}{\textbf{Method}} & \multirow{2}{*}{\textbf{Size}} & \multicolumn{5}{c|}{\textbf{MR-RefCOCO/+/g}} & \multicolumn{4}{c}{\textbf{MR-PACO}} \\
        ~ & ~ & \textbf{Round\#2} & \textbf{Round\#3} & \textbf{Round\#4} & \textbf{Round\#5} & \textbf{Round\#6} & \textbf{Round\#2} & \textbf{Round\#3} & \textbf{Round\#4} & \textbf{Round\#5}\\
        \hline
LISA~\cite{lisa} & 7B & 57.8 & 54.1 & 55.4 & 52.3 & 48.7 & 15.5 & 21.3 & 18.7 & 20.5 \\
GLaMM~\cite{glamm} & 7B & 59.3 & 59.2 & 56.3 & 54.2 & 52.7 & -- & -- & -- & -- \\
SegLLM ~\cite{segllm}& 7B & 79.7 & 78.7 & 76.6 & 74.6 & 72.3 & 49.7 & 40.9 & 39.4 & 41.9 \\
\rowcolor{blue!5}\textbf{Qwen25VL-SAMTok} & 3B & 83.3 & 79.9 & 80.1 & 77.1 & 77.8 & 55.0 & \underline{49.2} & 46.4 & 51.7 \\
\rowcolor{blue!5}\textbf{Qwen3VL-SAMTok} & 4B & \textbf{86.7} & \textbf{83.6} & \textbf{84.4} & \underline{78.5} & \textbf{85.3} & \textbf{58.7} & \textbf{51.1} & \textbf{50.8} & \textbf{54.0}\\
\rowcolor{blue!5}\textbf{Qwen25VL-SAMTok} & 7B & \underline{86.2} & \underline{83.2} & \underline{83.8} & \textbf{80.4} & \underline{84.9} & \underline{58.4} & 48.8 & \underline{48.2} & \underline{52.9}\\
    \bottomrule[1.5pt]
    \end{tabular}
    }
\end{table*}

\section{Experiment}
\label{sec:exp}

For implementation details, please refer to Appendix Sec. B. \textbf{For additional model experiments, please refer to Appendix Sec. C. For ablation studies, please refer to Appendix Sec. D.} For further visualization results, please refer to Appendix Sec. E.

\subsection{Experiment Setup}
\label{sec:exp-setup}

\begin{table*}[t!]
    \centering
    \caption{Results on text-to-mask task (GRES). ``(ft)''  indicates models further finetuned on GRES after mixed training. The best results are \textbf{bold} and the second-best results are \underline{underlined} (excluding our RL results). ``(rl)''  indicates models further trained with GRPO after mixed training.}
    \label{tab:exp-gres}
    \setlength{\tabcolsep}{5pt}
    \resizebox{0.95\textwidth}{!}{
    \begin{tabular}{l | c | c c c | c c c | c c c | c c c}
    \toprule[1.5pt]
        \multirow{2}{*}{\textbf{Method}} & \multirow{2}{*}{\textbf{Size}} & \multicolumn{3}{c|}{\textbf{Val}} & \multicolumn{3}{c|}{\textbf{Test A}} & \multicolumn{3}{c|}{\textbf{Test B}} & \multicolumn{3}{c}{\textbf{Avg.}}\\
        ~ & ~ & \textbf{gIoU} & \textbf{cIoU} & \textbf{N-acc} & \textbf{gIoU} & \textbf{cIoU} & \textbf{N-acc} & \textbf{gIoU} & \textbf{cIoU} & \textbf{N-acc} & \textbf{gIoU} & \textbf{cIoU} & \textbf{N-acc} \\
        \hline
LISA~\cite{lisa} & 7B & 61.6 & 61.8 & 54.7 & 66.3 & 68.5 & 50.0 & 58.8 & 60.6 & 51.9 & 62.2 & 63.6 & 52.2 \\
SAM4MLLM~\cite{chen2024sam4mllm} & 8B & 71.9 & 67.8 & 66.1 & 74.2 & 72.2 & 63.9 & 65.3 & 63.4 & 60.0 & 70.5 & 67.8 & 63.3 \\
MLLMSeg~\cite{wang2025MLLMSeg} & 8B & 75.1 & 71.6 & 73.2 & 77.0 & \underline{76.9} & 72.4 & 69.7 & 68.5 & 65.5 & \underline{73.9} & \underline{72.3} & 70.4\\
HiMTok~\cite{himtok} & 8B & 72.1 & 70.4 & -- & 73.5 & 74.9 & -- & \underline{71.7} & \textbf{72.0} & -- & 72.4 & \textbf{72.4} & -- \\
ARGenSeg~\cite{argenseg} & 8B & 74.7 & \textbf{72.2} & -- & 73.7 & 73.6 & -- & \textbf{72.4} & \underline{70.4} & -- & 73.6 & 72.1 & -- \\
\rowcolor{blue!5}\textbf{Qwen25VL-SAMTok} & 3B & 70.5 & 68.2 & 60.5 & 73.8 & 73.0 & 58.8 & 65.9 & 65.4 & 55.3 & 70.1 & 68.9 & 58.2 \\
\rowcolor{blue!5}\textbf{Qwen25VL-SAMTok (ft)} & 3B & \underline{77.1} & 71.2 & \underline{78.0} & 77.4 & 75.6 & 73.6 & 68.3 & 66.5 & \underline{67.2} & 74.3 & 71.1 & \underline{72.9} \\
\rowcolor{blue!5}\textbf{Qwen3VL-SAMTok} & 4B & 74.7 & 71.0 & 68.8 & 76.9 & 75.8 & 66.2 & 69.1 & 68.3 & 60.8 & 73.6 & 71.7 & 65.3\\ 
\rowcolor{blue!5}\textbf{Qwen25VL-SAMTok} & 7B & 74.4 & 71.4 & 67.5 & \underline{77.8} & 72.9 & \textbf{77.4} & 68.5 & 68.2 & 59.8 & 73.6 & 70.8 & 68.2 \\
\rowcolor{blue!5}\textbf{Qwen25VL-SAMTok (ft)} & 7B & \textbf{78.0} & \underline{72.1} & \textbf{79.5} & \textbf{78.5} & \textbf{77.3} & \underline{75.9} & 69.6 & 67.7 & \textbf{68.6} & \textbf{75.4} & \textbf{72.4} & \textbf{74.7}\\
        \hline
\rowcolor{blue!5}\textbf{Qwen25VL-SAMTok (rl)} & 3B & 79.4 & 73.7 & 81.5 & 79.4 & 77.7 & 77.7 & 71.3 & 69.8 & 72.0 & 76.7 & 73.7 & 77.1 \\
    \bottomrule[1.5pt]
    \end{tabular}
    }
\end{table*}

\begin{table*}[t]
    \centering
\begin{minipage}[t]{0.55\textwidth}
    \centering
    \captionof{table}{Results on text-to-mask task (RES). The evaluation metric is cIoU.}
    \setlength{\tabcolsep}{3pt}
    \label{tab:exp-res}
    \resizebox{\linewidth}{!}{
    \begin{tabular}{l | c | c c c | c c c | c c}
    \toprule[1.5pt]
        \multirow{2}{*}{\textbf{Method}} & \multirow{2}{*}{\textbf{Size}} & \multicolumn{3}{c|}{\textbf{RefCOCO}} & \multicolumn{3}{c|}{\textbf{RefCOCO+}} & \multicolumn{2}{c}{\textbf{RefCOCOg}} \\
        ~ & ~ & \textbf{val} & \textbf{test A} & \textbf{test B} & \textbf{val} & \textbf{test A} & \textbf{test B} & \textbf{val} & \textbf{test} \\
        \hline
Sa2VA~\cite{sa2va} & 4B & \underline{82.4} & \underline{84.2} & 79.5 & 77.6 & 81.2 & 73.1 & \underline{79.7} & 80.4 \\
PaDT Pro~\cite{padt} & 3B & 81.3 & 81.5 & \textbf{82.2} & 77.6 & 79.4 & \underline{76.3} & 78.1 & 78.5 \\
UniPixel~\cite{liu2025unipixel}& 3B & 81.9 & 83.5 & 78.6 & 75.3 & 80.3 & 70.6 & 77.2 & 78.5 \\
\rowcolor{blue!5}\textbf{Qwen25VL-SAMTok} & 3B & \underline{82.4} & 83.9 & 79.9 & \underline{78.4} & \underline{81.8} & 74.8 & 79.0 & \underline{79.1} \\
\rowcolor{blue!5}\textbf{Qwen3VL-SAMTok} & 4B & \textbf{83.4} & \textbf{85.0} & \underline{82.1} & \textbf{80.2} & \textbf{83.4} & \textbf{76.6} & \textbf{80.7} & \textbf{81.0} \\
    \bottomrule[1.5pt]
    \end{tabular}
    }
\end{minipage}
\hfill
\begin{minipage}[t]{0.43\textwidth}
    \centering
    \captionof{table}{Zero-shot results on text-to-mask task (GroundingSuite). The evaluation metric is gIoU.}
    \setlength{\tabcolsep}{1.5pt}
    \label{tab:exp-gseval-mask}
    \resizebox{\linewidth}{!}{
    \begin{tabular}{l | c | c c c c c}
    \toprule[1.5pt]
        \textbf{Method} & \textbf{Size} & \textbf{Stuff} & \textbf{Part} & \textbf{Multi} & \textbf{Single} & \textbf{All}\\
        \hline
LISA~\cite{lisa} & 7B & 85.2 & 21.2 & \underline{71.5} & 42.8 & 57.6 \\
GLaMM~\cite{glamm} & 7B & \underline{86.9} & 16.5 & 70.4 & 42.1 & 57.2 \\
EVF-SAM~\cite{zhang2024evf-sam} & 1B & 85.1 & 23.1 & \textbf{72.1} & \underline{54.5} & \underline{62.6} \\
InstructSeg~\cite{wei2025instructseg} & 3B & 56.2 & \underline{24.2} & 66.8 & 51.3 & 52.5 \\
\rowcolor{blue!5}\textbf{Qwen25VL-SAMTok} & 3B & \textbf{90.2} & \textbf{40.8} & 62.5 & \textbf{63.6} & \textbf{67.8} \\
    \bottomrule[1.5pt]
    \end{tabular}
    }
\end{minipage}
\end{table*}

\begin{table*}[t!]
    \centering\small
    \caption{Results on grounding task (REC). We de-tokenize the mask words generated by Qwen25VL-SAMTok into 2D masks and then derive bounding boxes for evaluation. }
    \label{tab:exp-rec}
    \setlength{\tabcolsep}{3pt}
    \resizebox{0.99\textwidth}{!}{
    \begin{tabular}{l | c | c c c | c c c | c c}
    \toprule[1.5pt]
        \multirow{2}{*}{\textbf{Method}} & \multirow{2}{*}{\textbf{Size}} & \multicolumn{3}{c|}{\textbf{RefCOCO}} & \multicolumn{3}{c|}{\textbf{RefCOCO+}} & \multicolumn{2}{c}{\textbf{RefCOCOg}} \\
        ~ & ~ & \textbf{val} & \textbf{test A} & \textbf{test B} & \textbf{val} & \textbf{test A} & \textbf{test B} & \textbf{val} & \textbf{test} \\
        \hline
Qwen25VL~\cite{qwen25vl} & 3B & 89.1 & 91.7 & 84.0 & 82.4 & 88.0 & 74.1 & 85.2 & 85.7 \\
\rowcolor{blue!5}\textbf{Qwen25VL-SAMTok} & 3B & 92.7 {\color{ForestGreen}(+3.6)} & 94.6 {\color{ForestGreen}(+2.9)} & 89.7 {\color{ForestGreen}(+5.7)} & 88.2 {\color{ForestGreen}(+5.8)} & 92.2 {\color{ForestGreen}(+4.2)} & 84.4 {\color{ForestGreen}(+10.3)} & 89.9 {\color{ForestGreen}(+4.7)} & 89.6 {\color{ForestGreen}(+3.9)} \\
Qwen25VL~\cite{qwen25vl} & 7B & 90.0 & 92.5 & 85.4 & 84.2 & 89.1 & 76.9 & 87.2 & 87.2 \\
\rowcolor{blue!5}\textbf{Qwen25VL-SAMTok} & 7B & 93.0 {\color{ForestGreen}(+3.0)} & 95.5 {\color{ForestGreen}(+3.0)} & 90.5 {\color{ForestGreen}(+5.1)} & 88.6 {\color{ForestGreen}(+4.4)} & 93.2 {\color{ForestGreen}(+4.1)} & 84.4 {\color{ForestGreen}(+7.5)} & 90.8 {\color{ForestGreen}(+3.6)} & 91.2 {\color{ForestGreen}(+4.0)} \\
    \bottomrule[1.5pt]
    \end{tabular}
    }
\end{table*}

\noindent\textbf{Training Data.}
Our training process consists of three stages: (1) SAMTok training, (2) MLLM supervised finetuning, and (3) reinforcement learning.

\noindent
\textbf{Tokenizer training data.}
To endow SAMTok with powerful mask compression and reconstruction capabilities, we collect diverse mask data~\cite{sam,coconut,entityseg,pixelweb,ade20k,cityscapes,yang2023lisa++,sam2,mevis,revos,refsam}.
As shown in Fig.~\ref{fig:dataset}, these mask data encompass various scenarios, including indoor scenes, outdoor environments, and website user interfaces, spanning multiple granularities, including part-level, object-level, entity-level, and semantic-level annotations, totaling 209M masks.

\noindent
\textbf{Supervised finetuning data.}
To enable the MLLM to develop a deep understanding and generation capability for the discrete mask tokens produced by SAMTok, we collect a large-scale dataset of interleaved mask-text data for supervised fine-tuning.
As illustrated in Fig.~\ref{fig:dataset}, this supervised fine-tuning data comprises three components:
(1) mask generation data, including grounding~\cite{invig,visual7w,refgta}, referring segmentation~\cite{refcoco,refcoco_p_g,refclef,cops_ref,gres}, grounded conversation generation~\cite{flickr30k,psg,glamm}, instance segmentation~\cite{v3det,coconut}, and scene parsing generation~\cite{psg} data;
(2) region understanding data, including region captioning~\cite{dam, gar} and region-based question answering~\cite{dam, gar} data;
(3) collaborative mask generation and understanding data, such as multi-turn interactive reasoning segmentation~\cite{segllm}.
The supervised finetune dataset totals approximately 5M samples.
All samples’ masks are pre-tokenized using SAMTok and reformatted into dialogue-style text sequences before training.

\noindent
\textbf{RL data.}
To enhance the reasoning capability of MLLMs in mask generation tasks, we first generate 26k cold-start samples by prompting the Qwen3-VL-235B model to simulate chain-of-thought (CoT) reasoning.
Additionally, 8k and 41k challenging samples are selected from the general referring segmentation dataset~\cite{gres} and the grounded conversation generation dataset~\cite{glamm} for GRPO~\cite{grpo}, respectively.


\subsection{Main Results}
\label{sec:main_results}

\noindent
\textbf{Interleaved Text-mask Generation Tasks.} We evaluate our model’s interleaved text–mask generation capability on the grounded conversation generation (GCG)~\cite{glamm} benchmark.
The GCG benchmark requires the model to describe an image while simultaneously generating region masks corresponding to the mentioned phrases.
As shown in Tab.~\ref{tab:exp-gcg}, our method achieves state-of-the-art (SOTA) performance on this benchmark.
On the validation set, the new SOTA improves over the previous best by \textbf{+1.3\%} METEOR and \textbf{+5.5\%} CIDEr in captioning metrics, and by \textbf{+5.3\%} AP50, \textbf{+5.2\%} mIoU, and \textbf{+4.7\%} Recall in mask metrics.
On the test set, similar consistent gains are observed, including \textbf{+1.2\%} METEOR, \textbf{+5.5\%} CIDEr, \textbf{+5.3\%} AP50, \textbf{+4.9\%} mIoU, and \textbf{+4.2\%} Recall.
These comprehensive improvements across both textual and visual dimensions demonstrate that the proposed SAMTok enables more precise text–mask alignment, effectively bridging language and pixel-level representations.

\noindent\textbf{Multi-round interactive segmentation task.}
This task requires the model to reason about complex user intentions and segment objects in relation to previously identified entities across multiple interaction rounds. 
These interactions involve positional, interactional, and hierarchical relationships between objects, demanding the model to maintain long-term visual–linguistic consistency. 
We adopt MR-RefCOCO/+/g~\cite{segllm} and MR-PACO~\cite{segllm} as evaluation benchmarks, which test the model’s ability to both \textit{embed masks in the input} and \textit{generate masks in the output}.
Our proposed SAMTok provides exactly such a unified \textit{mask-token interface} for MLLMs, enabling fine-grained spatial reasoning and compositional mask generation. As shown in Tab.~\ref{tab:exp-segllm}, our model achieves new SOTA results on both benchmarks. On MR-RefCOCO/+/g benchmark, the new SOTA surpasses the previous best by an average of \textbf{+7.7\%} across all interaction rounds, while on the MR-PACO benchmark, it yields an even larger average improvement of \textbf{+10.7\%}. 
These substantial gains demonstrate that our approach not only reasons about inter-object relations but also effectively models hierarchical part–whole relationships within objects, maintaining consistent segmentation quality through multi-turn interactions.

\noindent\textbf{Text-to-mask task.} We adopt the GRES~\cite{gres}, RefCOCO/+/g~\cite{refcoco, refcoco_p_g}, and GroundingSuite~\cite{groundingsuite} as our evaluation benchmarks. 
As shown in Tab.~\ref{tab:exp-gres}, although our model is trained solely with the \textit{next-token prediction loss}, it still outperforms existing methods that rely on task-specific loss on most metrics.
Specifically, our approach surpasses the previous best by \textbf{+1.5\%} in average gIoU across all three splits and achieves comparable performance in average cIoU, while improving average N-acc by a substantial \textbf{+4.3\%}.
As shown in Tab.~\ref{tab:exp-res}, our method is also competitive on the RefCOCO/+/g benchmarks; in particular, among models with fewer than 4B parameters, it achieves a new SOTA. 
We further conduct zero-shot evaluation on GroundingSuite, as shown in Tab.~\ref{tab:exp-gseval-mask}. 
Our method demonstrates stronger zero-shot capability than other region-level MLLMs (67.8 vs. 62.6) that use task-specific losses. 
These results highlight the strong generalization ability of our model, demonstrating that explicit mask supervision is not strictly necessary for effective text-to-mask reasoning.

\begin{table}[t!]
    \centering\small
    \caption{Results on mask-to-text task (DLC-Bench). This table reports region captioning performance on DLC-Bench, which evaluates the ability to generate accurate textual descriptions conditioned on given region masks.}
    \label{tab:exp-dam}

    \setlength{\tabcolsep}{8pt}

    \begin{tabular}{l | c | c c c }
    \toprule[1.5pt]
        \textbf{Method} & \textbf{Size} & \textbf{Pos.} & \textbf{Neg.} & \textbf{Avg.} \\
        \hline
GPT-4o~\cite{gpt4o} & -- & 43.4 & 79.6 & 61.5 \\
o1~\cite{o1} & -- & \underline{46.3} & 78.8 & 62.5 \\ 
Claude 3.7 Sonnet~\cite{claude37} & -- & 21.8 & 50.4 & 36.1 \\
Gemini 2.5 Pro~\cite{gemini} & -- & 36.5 & 75.2 & 55.8 \\
Qwen2.5VL~\cite{qwen25vl} & 7B & 20.3 & 62.2 & 41.2 \\
RegionGPT~\cite{guo2024regiongpt} & 7B & 10.6 & 46.4 & 28.5 \\
OMG-LLaVA~\cite{zhang2024omg} & 7B & 5.6 & 32.6 & 19.1 \\
VP-SPHINX~\cite{mdpv} & 13B & 26.3 & 71.6 & 49.0 \\
DAM~\cite{dam} & 3B & \textbf{52.3} & \underline{82.2} & \textbf{67.3} \\
\rowcolor{blue!5}\textbf{Qwen3VL-SAMTok} & 4B & 46.1 & \textbf{85.2} & \underline{65.6} \\
    \bottomrule[1.5pt]
    \end{tabular}
    
\end{table}

\noindent\textbf{Mask-to-text task.} Region caption is a representative mask-to-text task that requires producing textual descriptions conditioned on given region masks. We adopt three benchmarks for evaluation: DLC-Bench~\cite{dam}, MDVP-Bench~\cite{mdpv}, and VideoRefer-D~\cite{videorefer}.
Without introducing any architectural modifications to the base model, our approach achieves performance comparable to that of the expert model DAM~\cite{dam} on the DLC-Bench (65.6 vs.\ 67.3), as shown in Tab.~\ref{tab:exp-dam}.
In contrast, general MLLMs such as Qwen2.5VL-7B~\cite{qwen25vl}, despite being provided with richer prior information (e.g., category names), attain a much lower score of 41.2.
This proves our proposed SAMTok enables more precise and unambiguous region grounding for text generation.
We also conduct zero-shot evaluations on the MDVP-Bench and VideoRefer-D.
As shown in Tab.~\ref{tab:exp-mdvp}, our method surpasses the expert model DAM in three of the four metrics on MDVP-Bench, demonstrating strong generalization to document and panel-style visual scenes.
Similarly, on the video region captioning benchmark VideoRefer-D (Tab.~\ref{tab:exp-videorefer-d}), our approach also achieves competitive performance compared to other region-level MLLMs.

\begin{table}[t!]
    \centering\small
    \caption{Zero-shot region captioning results on MDVP-Bench across four different visual scene types: natural images, OCR-heavy documents, multi-panel layouts, and screenshots. The benchmark evaluates generalization to complex document-style and panel-based visual inputs without task-specific fine-tuning.}
    \label{tab:exp-mdvp}
    \setlength{\tabcolsep}{3pt}
    \resizebox{0.6\linewidth}{!}{
    \begin{tabular}{l | c | c c c c}
    \toprule[1.5pt]
        \textbf{Method} & \textbf{Size} & \textbf{Natural} & \textbf{OCR} & \textbf{Multi-Panel} & \textbf{Screenshot}\\
        \hline
Osprey~\cite{osprey} & 7B & \underline{107.7} & 99.4 & 70.0 & 81.3 \\
PAM~\cite{lin2025PAM} & 3B & 71.4 & 94.3 & \underline{86.8} & \underline{84.5} \\
DAM~\cite{dam} & 3B & 87.0 & \textbf{127.7} & 79.4 & 76.4 \\
\rowcolor{blue!5}\textbf{Qwen3VL-SAMTok} & 4B & \textbf{145.0} & \underline{127.3} & \textbf{145.4} & \textbf{109.6} \\
    \bottomrule[1.5pt]
    \end{tabular}
    }
\end{table}

\begin{table}[t!]
    \centering\small
    \caption{Zero-shot region captioning results on VideoRefer-D benchmark, evaluated under the single-frame setting. Scores are reported for the overall average (Avg.) and four sub-categories: subject correspondence (SC), appearance description (AD), temporal description (TD), and hallucination detection (HD).}
    \label{tab:exp-videorefer-d}
    \setlength{\tabcolsep}{6pt}
    \resizebox{0.6\linewidth}{!}{
    \begin{tabular}{l | c | c c c c c}
    \toprule[1.5pt]
        \textbf{Method} & \textbf{Size} & \textbf{Avg.} & \textbf{SC} & \textbf{AD} & \textbf{TD} & \textbf{HD}\\
        \hline
GPT-4o~\cite{gpt4o} & -- & \textbf{2.95} & \underline{3.34} & \underline{2.96} & \textbf{3.01} & 2.50 \\
Elysium~\cite{wang2024Elysium} & 7B & 1.57 & 2.35 & 0.30 & 0.02 & \textbf{3.59} \\
Ferret~\cite{ferret2} & 7B & 2.18 & 3.08 & 2.01 & \underline{1.54} & 2.14 \\
Osprey~\cite{osprey} & 7B & 2.34 & 3.19 & 2.16 & \underline{1.54} & 2.45 \\
\rowcolor{blue!5}\textbf{Qwen3VL-SAMTok} & 4B & \underline{2.88} & \textbf{4.48} & \textbf{3.14} & 1.28 & \underline{2.65}\\
    \bottomrule[1.5pt]
    \end{tabular}
    }
\end{table}

\noindent\textbf{RL results on GRES and GCG.}
We evaluate the thinking RL setting on the GRES~\cite{gres} benchmark and the non-thinking RL setting on the GCG~\cite{glamm} benchmark.
As shown in Tab.~\ref{tab:exp-gres}, applying \textit{purely textual rewards} leads to substantial performance improvements on GRES.
 Across all three splits, our model achieves an average gain of \textbf{+6.8\%} gIoU, \textbf{+4.9\%} cIoU, and \textbf{+18.9\%} N-acc.
 Similarly, as shown in Tab.~\ref{tab:exp-gcg}, on the GCG benchmark, the model exhibits consistent improvements across mask metrics, with \textbf{+4.5\%} AP50, \textbf{+2.0\%} mIoU, and \textbf{+6.6\%} Recall on average. Since we did not use any rewards that evaluate caption quality, the two caption metrics decreased accordingly.
To the best of our knowledge, this is the \textbf{first successful attempt} to optimize mask generation performance using \textit{text-only reward signals}.
 These results further verify that the proposed SAMTok enables tighter text–mask alignment and allows MLLMs to benefit from language-based reinforcement signals without direct pixel-level supervision.

 \noindent\textbf{Visual Grounding.}
To assess whether the mask-token interface provided by SAMTok outperforms the native text-box interface of MLLMs on grounding tasks, we also conduct experiments on the RefCOCO, RefCOCO+, and RefCOCOg benchmarks.
The results are shown in Tab.~\ref{tab:exp-rec}. 
Specifically, we de-tokenize mask words into 2D masks and then derive bounding boxes for evaluation.
Across both 3B and 7B model sizes, SAMTok yields substantial accuracy improvements while preserving the same natural-language interaction capabilities as the native text-box interface. 
For the 3D model
This further supports our motivation: the new mask representation (SAMTok) outperforms the point format for visual grounding.

\begin{figure}[t]
\centering
\includegraphics[width=0.99\textwidth]{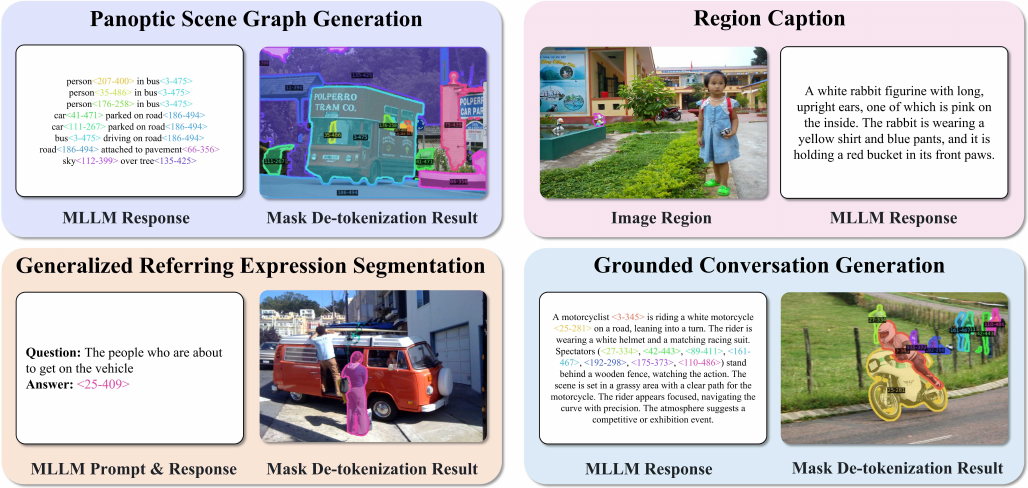}
\captionof{figure}{Qualitative examples of SAMTok on diverse downstream tasks, including panoptic scene graph generation (top-left), generalized referring expression segmentation (bottom-left), region captioning (top-right), and grounded conversation generation (bottom-right).}
\label{fig:task_examples}
\end{figure}

\noindent\textbf{Visualizations.}
Fig.~\ref{fig:task_examples} provides qualitative examples across multiple downstream tasks, including panoptic scene graph generation (PSG), generalized referring expression segmentation (GRES), region captioning, and grounded conversation generation (GCG). 
The visualizations highlight the effectiveness of SAMTok’s unified mask interface across diverse scenarios, enabling joint generation of structured relations and pixel-level masks, robust handling of complex referential language, accurate region-level descriptions, and precise localization of multiple phrases within long sentences.

\section{Related Work}
\label{sec:related_work}

\noindent\textbf{Mask Understanding in MLLMs.} In multimodal large language models (MLLMs)~\cite{qwen25vl,internvl35,cogvlm2,glm45v,gpt4o,o1,claude37}, accurately referring to and understanding spatial locations in images is crucial for fine-grained vision-language interaction. 
Recently, many works~\cite{som,vipllava,mmvm, qwenvl,qwen2vl,qwen25vl,internvl,internvl25,internvl3,dam,pixelrefer,gar,glamm,osprey} have focused on designing concise and effective region-level input representations to characterize users' spatial inputs.
These methods can be categorized into three approaches. (1) Visual prompts on images~\cite{som,vipllava,mmvm}, which directly highlight target regions to indicate the region of interest (ROI). This approach is intuitive and precise, but may alter the original image content.
(2) Textual coordinates~\cite{qwenvl,qwen2vl,qwen25vl,internvl,internvl25,internvl3,wang2025vgr,wang2025traceable,meng2025openo3video}, which specify regions via text-form 2D points/bounding boxes. While being most aligned with natural conversational interfaces, this approach often poses significant challenges for MLLMs in precisely identifying which image regions the coordinates refer to.
(3) ROI features~\cite{dam,pixelrefer,gar,glamm,osprey}, which either encode the region mask into image features or extract ROI features from feature maps using the region mask. 
These methods typically require specialized module designs, integrated architectures, and complex pipelines, resulting in limited generalizability and scalability.
Unlike the above approaches, our proposed SAMTok can compress input regions into representations of 2 special text tokens, thereby overcoming their deficiencies, including not affecting image content, enabling precise and efficient representation of regions, and being completely decoupled from the MLLM.

\noindent\textbf{Mask Generation in MLLMs.}
Spatial localization capability is crucial for MLLMs, as it constitutes a fundamental component of their understanding of the physical world.
Representing region-level outputs is a key challenge, as it not only determines the upper bound of spatial localization performance but also affects whether relevant data can be leveraged to efficiently instill spatial understanding capabilities into MLLMs.
Existing approaches can be categorized into three types.
(1) Directly outputting 2D points~\cite{guo2025seed1}, bounding boxes~\cite{qwen25vl, internvl3, guo2025seed1}, or polygonal contours~\cite{ferret2,wang2023visionllm,vectorllm} in textual form, which is straightforward but struggles to achieve strong performance since there exists a critical gap between textual cross-entropy and continuous localization.
(2) Aggregating segmentation information via special tokens~\cite{lisa, sa2va, hyperseg, xsam, zhang2024omg, padt, zhang2025pixel} and decoding the tokens into 2D masks with a dedicated segmentation model~\cite{sam,mask2former}, which requires joint training of the MLLM with an additional segmentation decoder and relies on segmentation loss optimization.
%
(3) treating a region mask as an image~\cite{himtok,alto,argenseg,lan2024text4seg} and autoregressively generating mask images, which can yield high-precision segmentation but incur substantial computational costs.
Compared to the above technical approaches, the core advantages of our SAMTok lie in its compact and efficient representation, high-fidelity mask reconstruction, and improved text-mask alignment, thereby substantially enhancing MLLMs' capability to understand and generate spatial regions while preserving natural-language interaction.

\noindent\textbf{Reinforcement Learning in Pixel-wise MLLMs.}
Reinforcement Learning~\cite{rlhf,dpo,ppo,grpo} (RL) has become a crucial post-training step for MLLMs, significantly enhancing the model's capability ceiling.
Some works~\cite{ma2025rethinking,visionr1,visionsr1,wang2025traceable} have achieved tremendous success with RL on mathematics, coding, and multiple-choice VQA by measuring rewards through simple and effective answer matching. 
In the RL process for pixel-wise MLLMs~\cite{segzero,visionreasoner,vlmr1,lens,visualrft}, reward is typically assessed via box IoU or mask IoU, which requires involving a mask decoder to convert text boxes or hidden states into 2D masks.
Thanks to SAMTok's efficient textual region mask representation, a purely textual answer-matching reward for region mask generation becomes feasible, computed by checking whether the predicted answer contains the target region's textual special tokens.
Our approach removes the need for any de-tokenization process and does not rely on external tools or auxiliary models for correctness verification.

\section{Conclusion}
\label{sec:conclusion}

This paper introduces SAMTok, a SAM-based mask tokenizer that represents any region mask using only two discrete tokens, enabling MLLMs to both generate and understand image-region masks within a unified textual interface. 
By converting region masks into special textual tokens, SAMTok reformulates mask understanding and generation as standard next-token prediction, removing the need for task-specific architectures or loss functions. 
Furthermore, we make the first successful attempt to optimize mask generation performance using purely text-based reward signals. 
Extensive experiments demonstrate that SAMTok significantly enhances pixel-wise understanding and generation across diverse benchmarks. 
We hope SAMTok points toward a scalable, unified, and language-native paradigm for pixel-wise visual reasoning in future multimodal systems.

\noindent
\textbf{Future Work Discussion.}
SAMTok is trained to reconstruct 2D image region masks and, therefore, lacks the ability to reconstruct region masks in video. 
In addition, SAMTok can only handle mask representation, where more visual entities, such as points, lines, and boxes, should also be included. 
Adding more visual entities enables more flexible interaction between VLMs and human inputs, which poses more chanlleges into tokener designs.
Lastly, we will also explore applications of SAMTok in video tasks, general VQA tasks, and image generation and editing.

\noindent
\textbf{Impact.} Our work explores a new mask tokenization design, which rethinks current pixel-wise multimodal large language model designs. 
Our work bridges regional-level understanding and reasoning through a new perspective, making pixel-LLM training as easy as MLLM (LMM) training.


\clearpage
\beginappendix
\appendix

\section{Overview}
\label{sec:appendix-overview}

In this file, we present additional experimental results due to space constraints in the main paper.
Here are the details:
\begin{itemize}
    \item In \S~\ref{sec:appendix-setup}, we provide more implementation details for both the tokenizer and the multi-modal large language models.
    \item In \S~\ref{sec:appendix-model-exp}, we provide more experimental results.
    \item In \S~\ref{sec:appendix-ablation}, we present ablation studies on tokenizer designs and their effectiveness.
    \item In \S~\ref{sec:appendix-vis}, we provide more visualizations.
\end{itemize}

\section{Implementation Details}
\label{sec:appendix-setup}

We initialize the encoder and decoder of SAMTok with the pretrained weights of SAM 2.1~\cite{sam2} (Large), while the quantizer's codebook is randomly initialized. 
During training, the parameters of the SAM image encoder and SAM prompt encoder are frozen, whereas the SAM decoder is trainable. Unless otherwise specified, the SAMTok codebook size is set to 256, and the number of quantization steps is 2, with the two codebooks non-shared. 
We train SAMTok using Xtuner~\cite{xtuner} and the AdamW optimizer~\cite{AdamW}, with a global batch size of 1024, an initial learning rate of 4e-5, and a cosine decay schedule~\cite{loshchilov2016sgdr}.
We employ Qwen-VL series~\cite{qwen25vl} models as the base model for the main experiments. 
The only modification made to the base models is the addition of mask tokens in the vocabulary. By default, we introduce 512 mask tokens, formatted as \texttt{<|mt\_0000|>} $\sim$ \texttt{<|mt\_0511|>}. 
Among them, \texttt{<|mt\_0000|>} $\sim$ \texttt{<|mt\_0255|>} correspond to the first-level codebook, and \texttt{<|mt\_0256|>} $\sim$ \texttt{<|mt\_0511|>} correspond to the second-level codebook.
In addition, we introduce two special tokens, \texttt{<|mt\_start|>} and \texttt{<|mt\_end|>}, to denote the start and end positions of mask token sequences. 
The embeddings of these newly added tokens are randomly initialized using the mean and variance statistics of the original token embeddings.
During supervised fine-tuning (SFT) and reinforcement learning (RL), we freeze the MLLM image encoder and fine-tune the projection layer and LLM. 
SFT is conducted using the official Qwen-VL~\cite{qwen2vl,qwen25vl} implementation, the AdamW optimizer~\cite{AdamW}, a global batch size of 256, and a learning rate of 2e-5 with a cosine decay schedule~\cite{loshchilov2016sgdr}. 
All training and evaluation experiments are performed on NVIDIA A100 GPUs (80 GB).
For RL, we employ the Easy-R1~\cite{easyr1} framework with the GRPO~\cite{grpo} algorithm, using a learning rate of 1e-6.

\section{Additional model experiments}
\label{sec:appendix-model-exp}

We present more experiments for more vision-language tasks.

\noindent\textbf{Panoptic scene graph generation.}
We further evaluate our model’s interleaved text–mask generation capability beyond the GCG benchmark on the more challenging PSG benchmark~\cite{psg}.
This task requires predicting all subject–relation–object triplets in an image, where each subject and object includes its category label and 2D mask.
As shown in Tab.~\ref{tab:exp-psg}, our Qwen2.5VL-SAMTok-7B achieves performance comparable to expert models (\mbox{R@20 = 19.8 vs.\ 20.6}, \mbox{mR@20 = 15.4 vs.\ 14.8}).
This result indicates that the mask-token interface provided by SAMTok enables effective task generalization by converting 2D masks into specialized word tokens.
Note that, our model \textbf{does not} need any task-specific designs, compared with previous expert models.

\begin{table}[t!]
    \centering
    \caption{Results on panoptic scene graph generation.}\vspace{-2mm}
    \label{tab:exp-psg}
    \setlength{\tabcolsep}{5pt}
    \resizebox{0.4\columnwidth}{!}{
    \begin{tabular}{l | c c}
    \toprule[1.5pt]
        \textbf{Method} & \textbf{R@20} & \textbf{mR@20}\\
        \hline
IMP~\cite{imp} & 16.5 & 6.5 \\
GPSNet~\cite{gpsnet} & 17.8 & 7.0 \\
MOTIF~\cite{motif} & \underline{20.0} & 9.1 \\
VCTree~\cite{vctree} & \textbf{20.6} & 9.7 \\
PSGFormer~\cite{psg} & 18.0 & \underline{14.8} \\
\rowcolor{blue!5}\textbf{Qwen25VL-SAMTok-7B} & 19.8 & \textbf{15.4} \\
    \bottomrule[1.5pt]
    \end{tabular}
    }
\end{table}

\begin{table*}[t!]
    \centering\small
    \caption{Effectiveness of unified mask-token interface for different MLLMs. For the GCG benchmark, we report the average of each metric across the val and test splits; for the GRES benchmark, we report the average across the val, testA, and testB splits.}
    \label{tab:exp-mllms}
    \setlength{\tabcolsep}{5pt}
    \resizebox{0.95\textwidth}{!}{
    \begin{tabular}{l | c | c c c c c | c c c | c c c}
    \toprule[1.5pt]
        \multirow{2}{*}{\textbf{Method}} & \multirow{2}{*}{\textbf{Size}} & \multicolumn{5}{c|}{\textbf{GCG}} & \multicolumn{3}{c|}{\textbf{GRES}} & \multicolumn{3}{c}{\textbf{DLC-Bench}} \\
        ~ & ~ & \textbf{METEOR} & \textbf{CIDEr} & \textbf{AP50} & \textbf{mIoU} & \textbf{Recall} & \textbf{gIoU} & \textbf{cIoU} & \textbf{N-acc} & \textbf{Pos.} & \textbf{Neg.} & \textbf{Avg.} \\
        \hline
\textbf{Qwen25VL-SAMTok} & 3B & 16.8 & 52.2 & 36.4 & 71.3 & 47.0 & 70.1 & 68.9 & 58.2 & 45.2 & 74.8 & 60.0 \\
\textbf{Qwen3VL-SAMTok} & 4B & 16.1 & 50.8 & 37.6 & 71.4 & 47.9 & 73.6 & 71.7 & 65.3 & 46.1 & 85.2 & 65.6 \\
\textbf{PLM~\cite{plm}-SAMTok} & 1B & 17.4 & 56.2 & 41.9 & 74.8 & 51.8 & 74.7 & 72.9 & 66.4 & 44.3 & 83.4 & 63.9 \\
    \bottomrule[1.5pt]
    \end{tabular}
    }
\end{table*}


\noindent\textbf{SAMTok integration across MLLMs.}
SAMTok and MLLM are decoupled: once SAMTok is trained, it can be deployed with any MLLM. 
To substantiate this, we train and evaluate on two types of MLLMs: (1) models with tile-based visual encoders (e.g., PerceptionLM~\cite{plm}), and (2) models with adaptive-resolution encoders (e.g., the Qwen-VL~\cite{qwen25vl} series). 
We use the same training data across settings, and report results in Tab.~\ref{tab:exp-mllms}. 
The mask-token interface provided by SAMTok works effectively across diverse MLLMs.

\section{Ablation Study}
\label{sec:appendix-ablation}

\noindent
\textbf{Set Up.} We evaluate SAMTok’s region-mask reconstruction capability on the EntitySeg~\cite{entityseg} validation set, which provides 23,754 region masks with high-quality annotations. 
We use mask IoU as the reconstruction accuracy metric (r-Acc). 
To assess the mask generation capability of an MLLM integrated with SAMTok, we adopt Qwen2.5-VL-3B~\cite{qwen25vl} as the base model and report the mean cIoU on the val splits of RefCOCO, RefCOCO+, and RefCOCOg in the RES setting as the generation accuracy metric (g-Acc).

\noindent\textbf{Quantization.}
We study three quantization strategies, including VQ~\cite{vqvae}, FSQ~\cite{fsq}, and RQ~\cite{rq}, as summarized in Tab.~\ref{tab:ablation-quant}.
When adopting standard VQ, a large codebook is essential, as region mask embeddings in images exhibit high diversity. 
Reducing the codebook size from 65,536 to 1,024 results in a significant drop in both reconstruction and generative accuracy. 
With a large codebook, FSQ improves codebook utilization, leading to better reconstruction and generation performance than standard VQ. 
RQ achieves comparable or even better performance while substantially reducing the codebook size. 
Specifically, using only two residual codebooks of size 1,024 each (i.e., 1024$\times$2), RQ attains higher reconstruction and generation accuracy than FSQ with a much larger codebook.

\noindent\textbf{Codebook size and quantization steps.} We further analyze the effect of codebook size and the number of residual quantization steps on reconstruction and generation performance, as shown in Tab.~\ref{tab:ablation-rq}. 
For the same codebook size, increasing the number of quantization steps (e.g., 1024$\times$4 vs.\ 1024$\times$2) yields higher-fidelity quantization and thus better reconstruction accuracy (0.75 vs.\ 0.70 in r-Acc). 
However, the exponentially expanded search space (1024$^4$ vs.\ 1024$^2$) makes it more difficult for the MLLM to learn to generate mask tokens effectively. 
In particular, in the dense mask setting, longer words incur much higher computational costs.
Therefore, we set the number of residual steps to 2 by default. 
Under this configuration, increasing the codebook size yields only a limited improvement in reconstruction accuracy but slightly enhances generation accuracy. 
We finally adopt the 256$\times$2 configuration as our default, since it offers a compact codebook while maintaining a substantial trade-off between reconstruction and generation performance.
We use this setting for further experiments, including SFT and RL processes.

\begin{table}[t!]
    \centering\small
    \caption{Ablation study on the quantization strategy. ``1024$\times$2'' means two residual codebooks with a size of 1024 each.}\vspace{-2mm}
    \label{tab:ablation-quant}
    \setlength{\tabcolsep}{8pt}
    \begin{tabular}{l | l | c c}
    \toprule[1.5pt]
        Quantization & Codebook Size & r-Acc & g-Acc \\
        \hline
        VQ~\cite{vqvae} & 1024 & 0.50 & 63.3 \\
        VQ~\cite{vqvae} & 65536 & 0.66 & 77.8 \\
        FSQ~\cite{fsq} & 65536 & \underline{0.69} & \underline{78.1} \\
        RQ~\cite{rq} & 1024$\times$2 & \textbf{0.70} & \textbf{78.3} \\
    \bottomrule[1.5pt]
    \end{tabular}
\end{table}

\begin{table}[t!]
    \centering\small
    \caption{Ablation study on the codebook size and quantization steps when using RQ~\cite{rq}.}\vspace{-2mm}
    \label{tab:ablation-rq}

    \setlength{\tabcolsep}{20pt}
    \begin{tabular}{l | l | c c}
    \toprule[1.5pt]
        Codebook Size & r-Acc & g-Acc \\
        \hline
        1024$\times$2 & 0.70 & \textbf{78.3} \\
        1024$\times$4 & \textbf{0.75} & 77.3 \\
        256$\times$4 & \underline{0.72} & 77.3 \\
        256$\times$2 & 0.70 & 77.6 \\
        512$\times$2 & 0.71 & \underline{77.8} \\
    \bottomrule[1.5pt]
    \end{tabular}
\end{table}

\section{Visualization}
\label{sec:appendix-vis}

In all visualizations in the main paper and the supplementary material, we represent mask words using quant-code pairs: for example, ``\texttt{<|mt\_0011|>}\texttt{<|mt\_0347|>}'' is denoted as ``\texttt{<11-347>}''.

\begin{figure*}[t]
\centering
\includegraphics[width=0.99\textwidth]{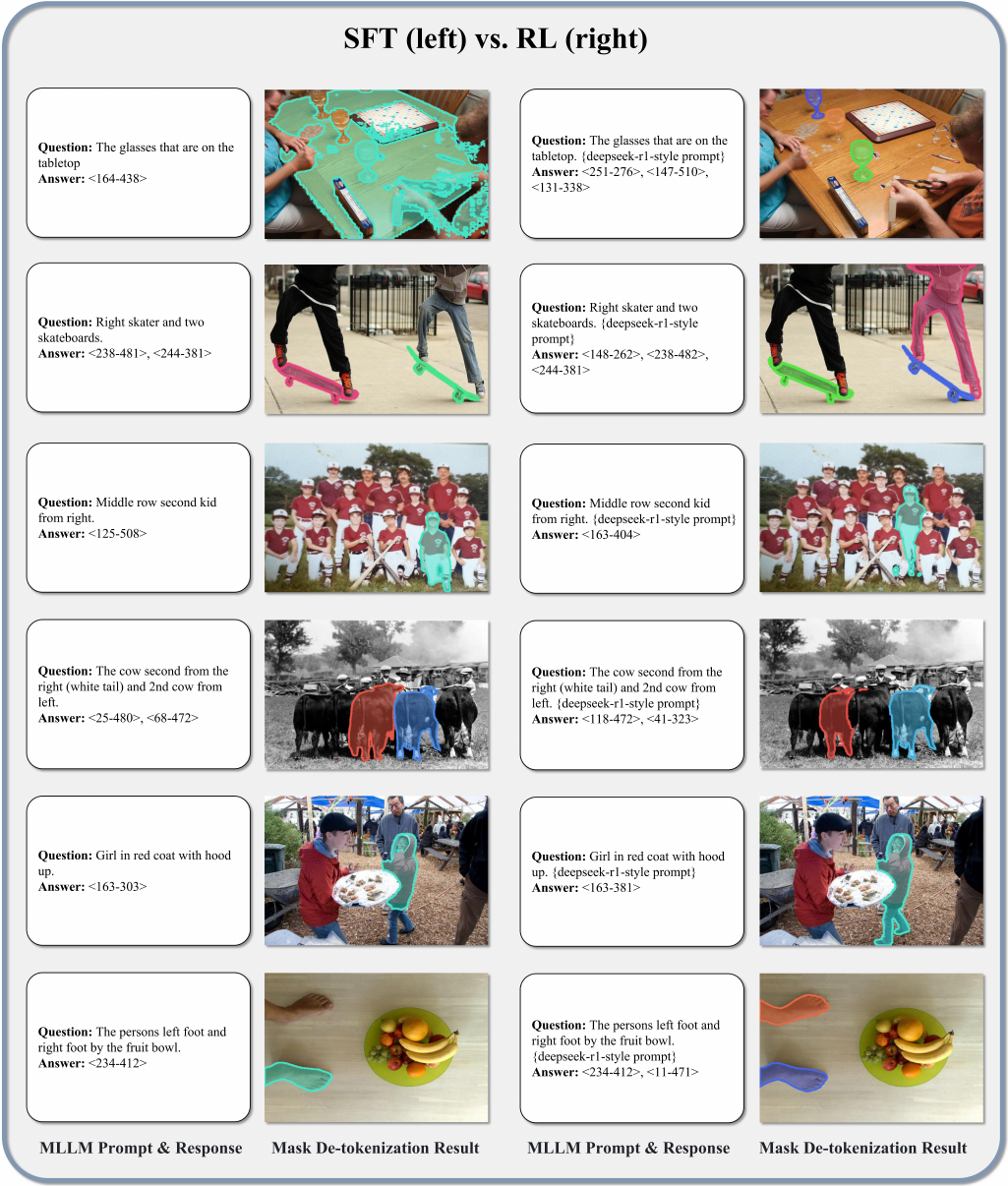}
\vspace{-2mm}\caption{\small SFT vs. RL. Examples are sampled from the GRES benchmark. RL finds more target objects, localizes relative positions better, and produces cleaner masks than SFT across diverse scenes.}
\label{fig:sft_vs_rl}
\end{figure*}

\noindent\textbf{SFT vs. RL.}
We visualize the improvements brought by RL over SFT in Fig.~\ref{fig:sft_vs_rl}. 
The gains manifest in three key aspects: (1) higher recall of targets in multi-object grounding scenarios; (2) more accurate localization for expressions involving relative positions and ordering; and (3) improved mask quality.

\begin{figure*}[t]
\centering
\includegraphics[width=0.99\textwidth]{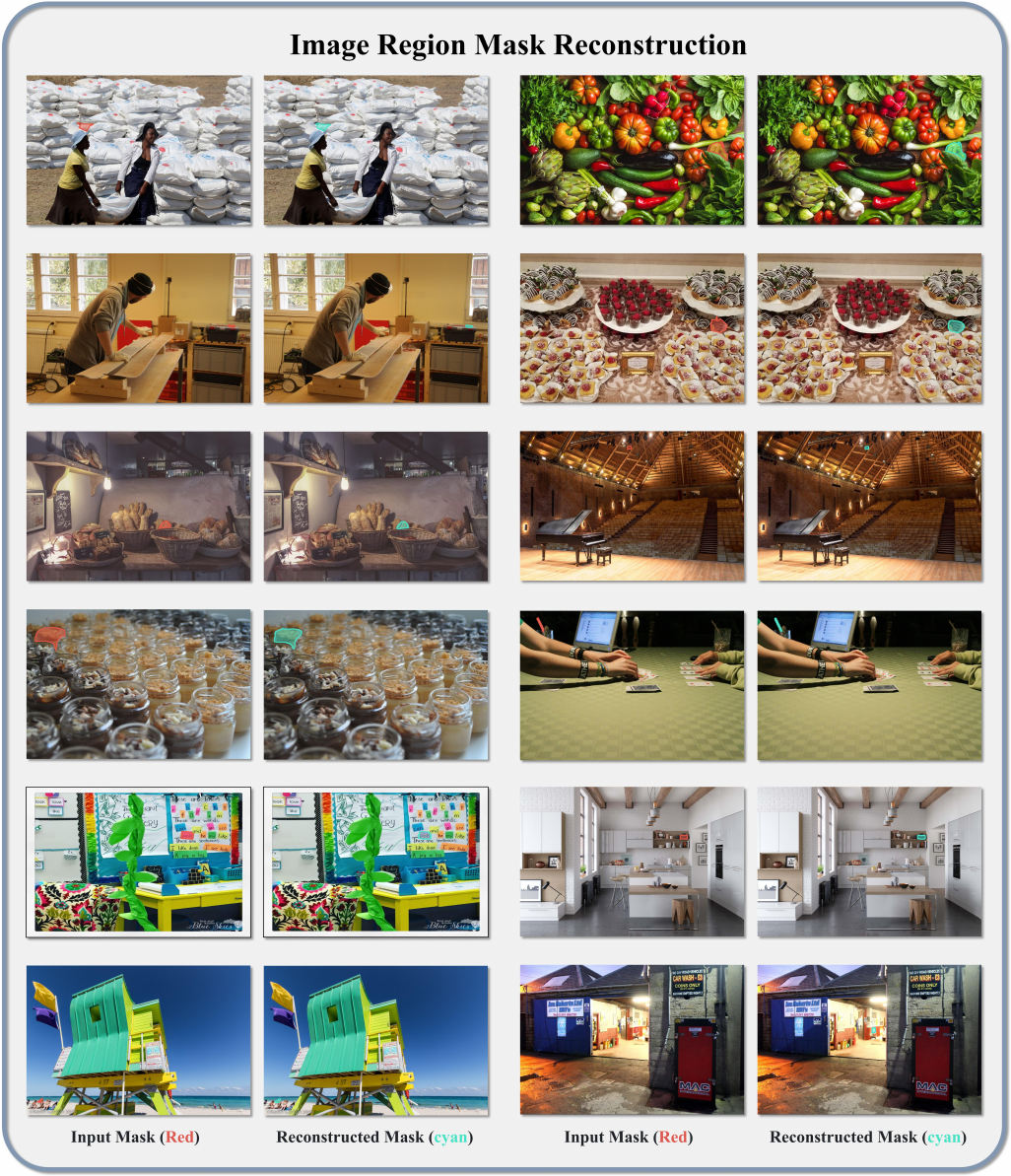}
\vspace{-2mm}\caption{\small Region mask reconstruction examples. For each region, the ground-truth mask is tokenized into two discrete codes, and SAMTok reconstructs the mask solely from the original image and the quantized mask tokens. SAMTok preserves fine structures for small, thin, or irregular objects even under challenging lighting or clutter. Since SAMTok is fully decoupled from the MLLM, its reconstruction quality remains stable regardless of downstream model training—unlike joint-training mask tokenizers that tend to collapse to elliptical or blurred masks.}
\label{fig:mask_recon_examples}
\end{figure*}

\noindent\textbf{Mask reconstruction.}
In Fig.~\ref{fig:mask_recon_examples}, we demonstrate SAMTok’s ability to reconstruct small objects across a variety of challenging scenarios. 
Since SAMTok and the MLLM are decoupled, the mask reconstruction capability is unaffected by any subsequent MLLM training. 
In contrast, other mask tokenizers~\cite{himtok,alto} require joint training with the MLLM, which ultimately leads to degraded mask reconstruction performance (with all masks reconstructed as ellipses). 
Thus, our method yields stronger mask reconstruction than these methods.
Such mask reconstruction generalization further leads to strong generalization in MLLMs.

\begin{figure*}[t]
\centering
\includegraphics[width=0.99\textwidth]{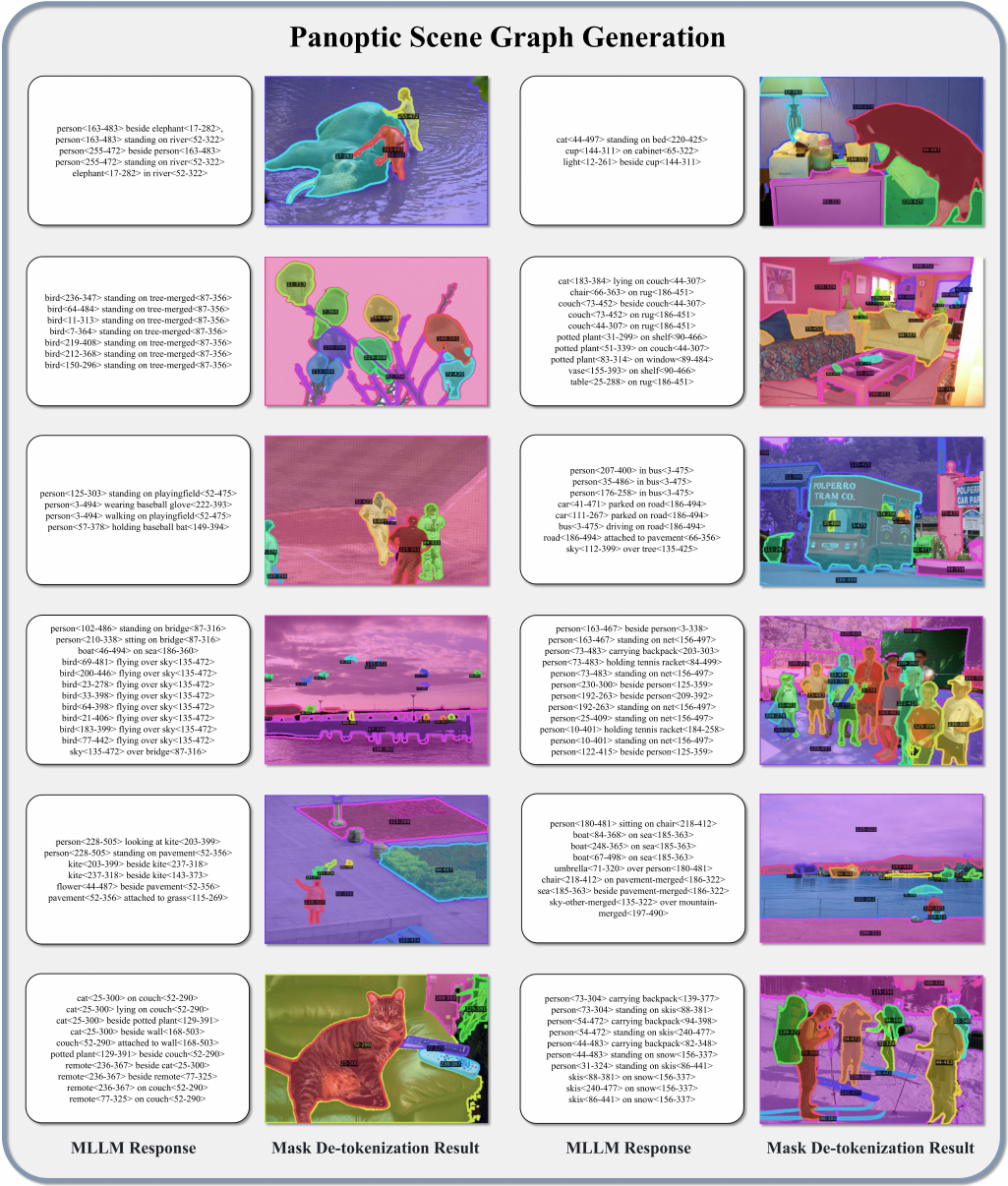}
\vspace{-2mm}\caption{\small Panoptic scene graph generation(PSG) examples. The model predicts subject–relation–object triplets where both subject and object categories are paired with their corresponding segmentation masks, represented through mask tokens. SAMTok’s interface allows the MLLM to generate consistent object masks and relational descriptions simultaneously, demonstrating strong alignment between textual predicates and pixel-grounded regions.}
\label{fig:psg_examples}
\end{figure*}

\noindent\textbf{PSG visualizations.}
Fig.~\ref{fig:psg_examples} shows PSG predictions. Each example contains subject–relation–object triplets, where both the subject and the object include their segmentation masks. Thanks to the unified mask-token interface, the MLLM can jointly generate structured relational descriptions and pixel-aligned masks, enabling dense, panoptic-level scene understanding.

\begin{figure*}[t]
\centering
\includegraphics[width=0.99\textwidth]{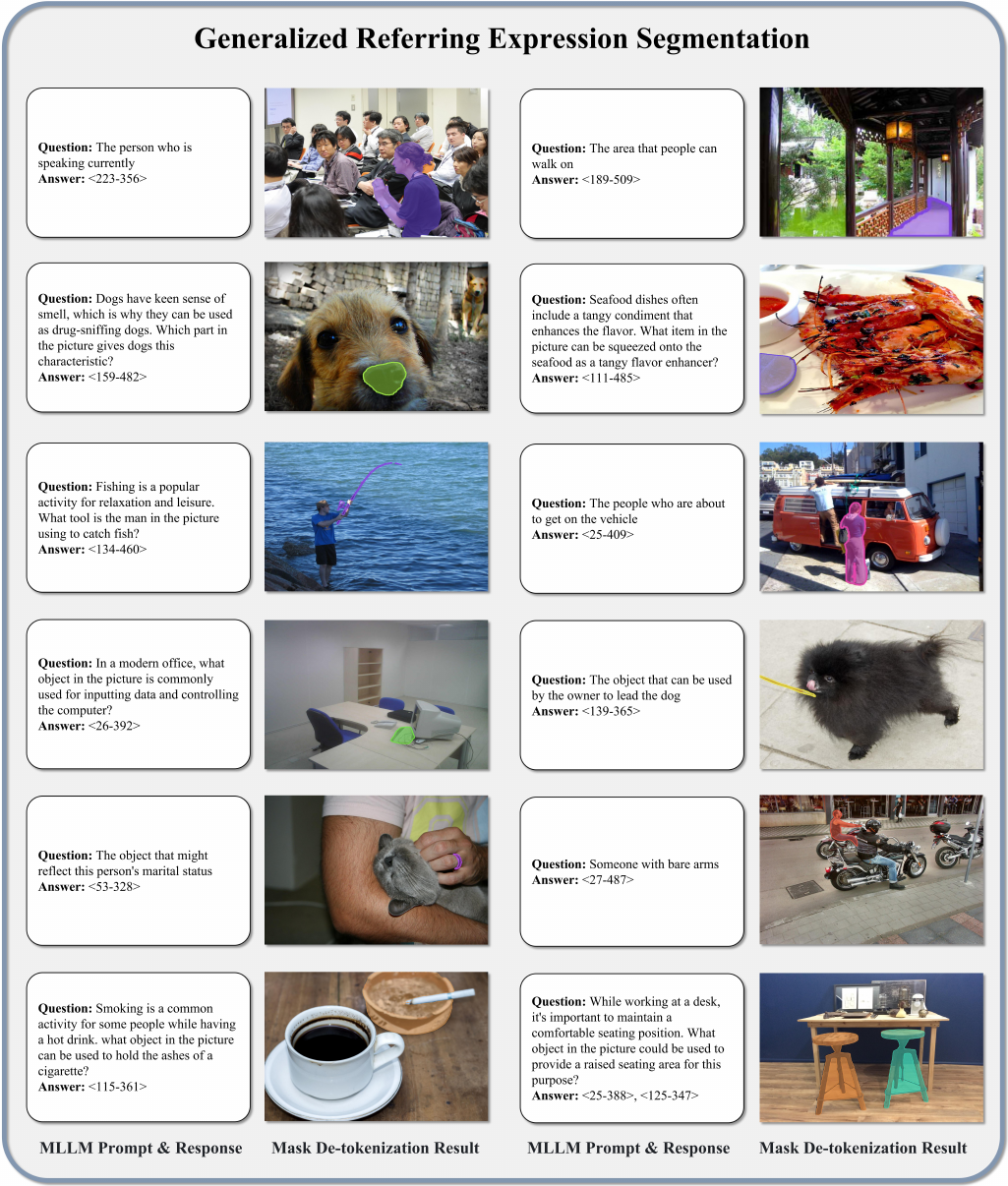}
\vspace{-2mm}\caption{\small GRES examples. Given a natural-language referring expression, the MLLM outputs two mask tokens that decode into the final segmentation mask. SAMTok enables precise grounding for expressions involving fine attributes, part-level targets, or contextual reasoning. The examples show robustness to ambiguous descriptions, occlusion, and multi-object scenes.}
\label{fig:gres_examples}
\end{figure*}

\noindent\textbf{GRES visulizations.}
Fig.~\ref{fig:gres_examples} illustrates GRES predictions. SAMTok allows the MLLM to resolve complex referring expressions, including attribute-dependent, context-dependent, and part-level descriptions. The generated masks accurately track the intended region even under heavy occlusion or clutter.

\begin{figure*}[t]
\centering
\includegraphics[width=0.99\textwidth]{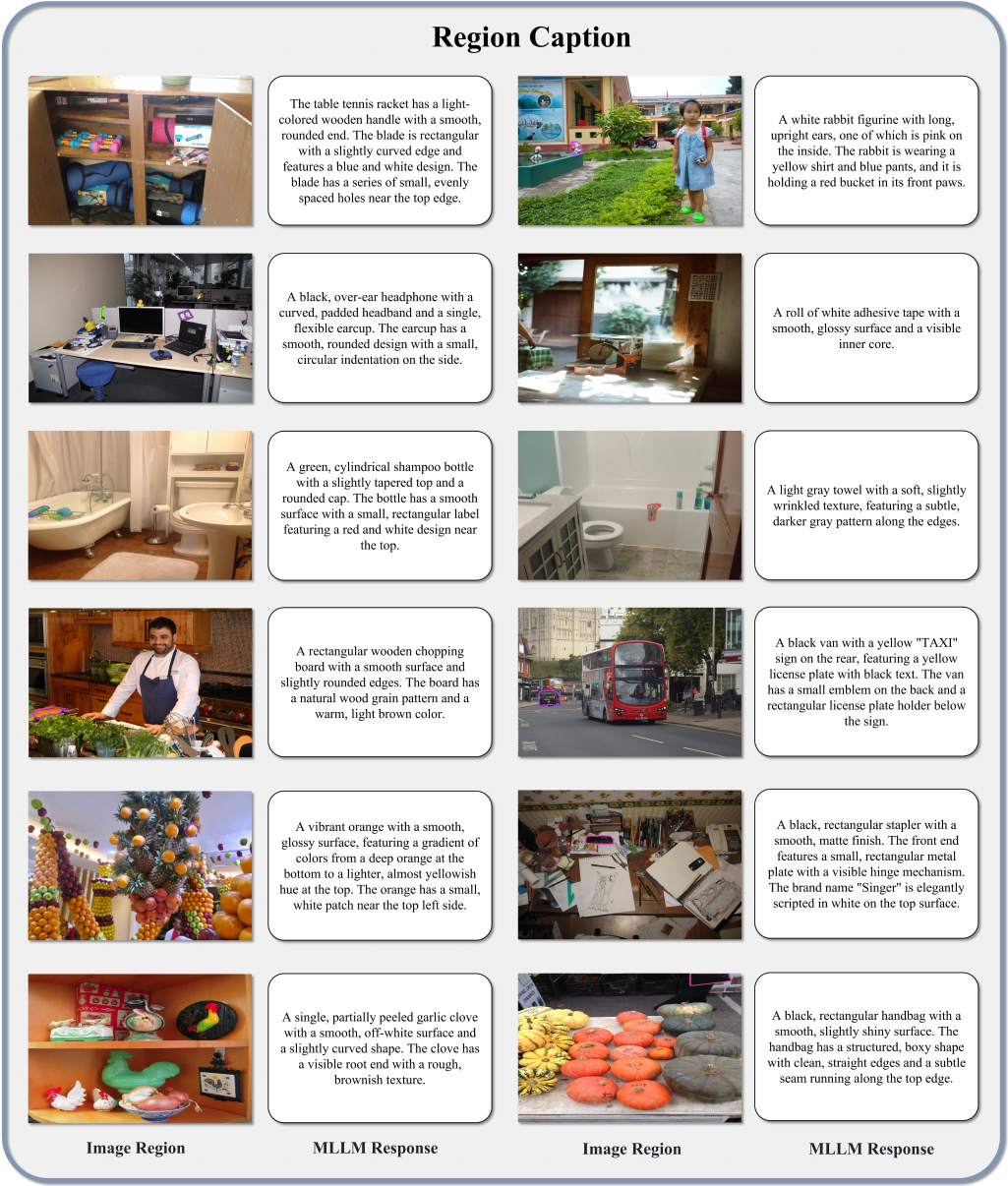}
\vspace{-2mm}\caption{\small Region Caption examples. Each visualization shows a region mask input (tokenized as two mask tokens) and the model’s generated description. SAMTok provides unambiguous spatial grounding, enabling the MLLM to generate accurate and context-aware region descriptions about attributes, roles, and interactions.}
\label{fig:dam_examples}
\end{figure*}

\noindent\textbf{Region caption visualizations.}
Fig.~\ref{fig:dam_examples} presents region captioning results. Given a region mask (tokenized into two special tokens), the MLLM produces detailed and contextually relevant descriptions. The compact mask tokens remove ambiguity inherent in bounding-box-based grounding and lead to more precise and consistent region-level captioning.

\begin{figure*}[t]
\centering
\includegraphics[width=0.99\textwidth]{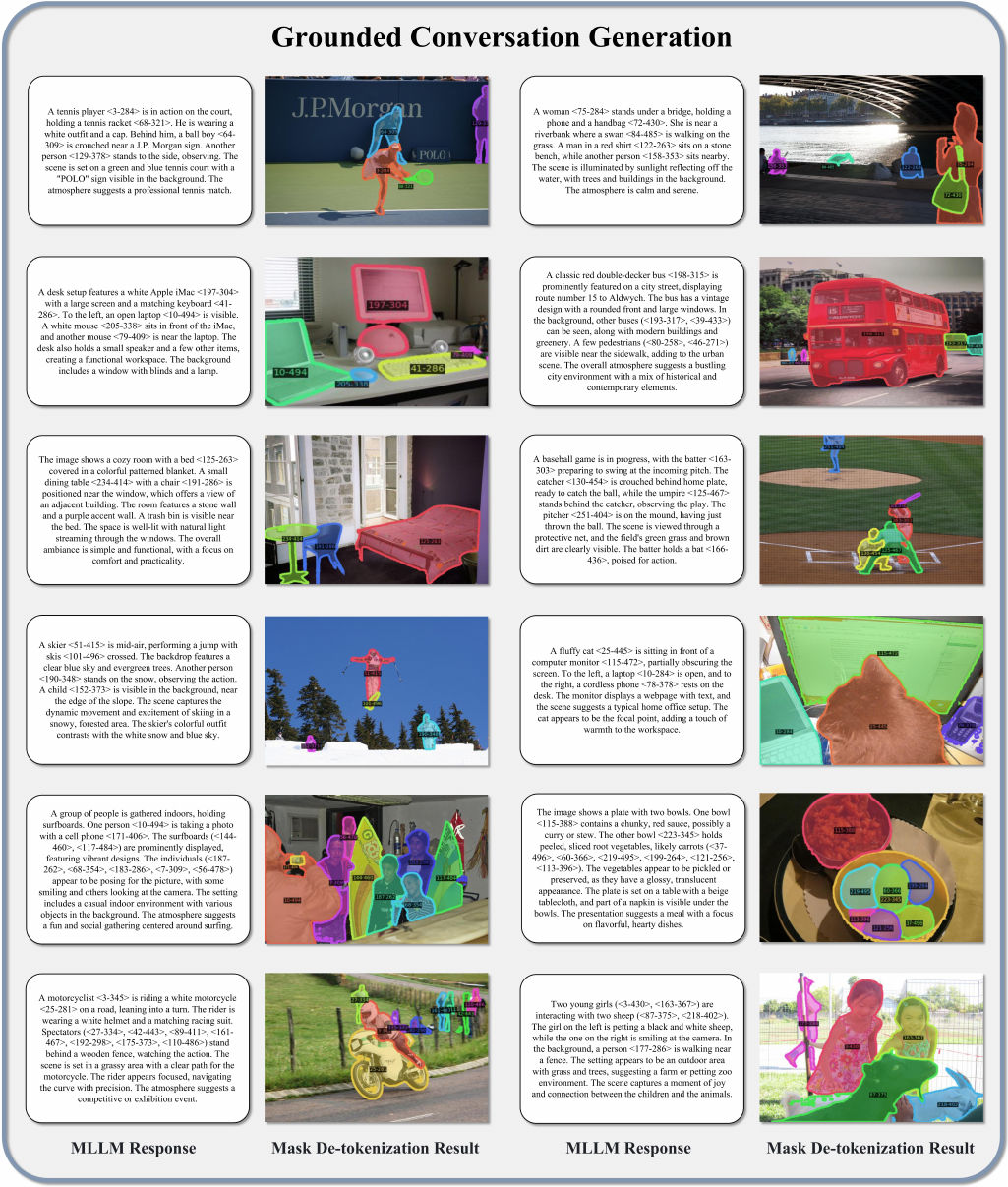}
\vspace{-2mm}\caption{\small GCG examples. The model simultaneously describes the scene and produces region masks for phrases mentioned in the caption. For each highlighted phrase, SAMTok decodes the predicted mask tokens into segmentation masks. SAMTok’s compact representation (two tokens per mask) enables efficient, aligned text–mask generation with consistent grounding across multiple phrases within long captions.}
\label{fig:gcg_examples}
\end{figure*}

\noindent\textbf{GCG visualizations.}
Fig.~\ref{fig:gcg_examples} shows examples where the model simultaneously generates textual captions and region masks for phrases mentioned in the narrative. SAMTok provides a lightweight and efficient mechanism for linking each phrase with a precise pixel region, enabling both high-quality caption generation and aligned mask prediction within one unified response.

\clearpage

\clearpage 

\clearpage 

\clearpage 

\clearpage 

\clearpage 

\clearpage
\bibliographystyle{plainnat}
\bibliography{bytedance/arxiv_bytedance}

\begin{thebibliography}{102}
\providecommand{\natexlab}[1]{#1}
\providecommand{\url}[1]{\texttt{#1}}
\expandafter\ifx\csname urlstyle\endcsname\relax
  \providecommand{\doi}[1]{doi: #1}\else
  \providecommand{\doi}{doi: \begingroup \urlstyle{rm}\Url}\fi

\bibitem[Anthropic(2025)]{claude37}
Claude Anthropic.
\newblock 3.7 sonnet and claude code, 2025.

\bibitem[Bai et~al.(2023)Bai, Bai, Yang, Wang, Tan, Wang, Lin, Zhou, and Zhou]{qwenvl}
Jinze Bai, Shuai Bai, Shusheng Yang, Shijie Wang, Sinan Tan, Peng Wang, Junyang Lin, Chang Zhou, and Jingren Zhou.
\newblock Qwen-vl: A versatile vision-language model for understanding, localization, text reading, and beyond.
\newblock \emph{arXiv preprint arXiv:2308.12966}, 2023.

\bibitem[Bai et~al.(2025)Bai, Chen, Liu, Wang, Ge, Song, Dang, Wang, Wang, Tang, Zhong, Zhu, Yang, Li, Wan, Wang, Ding, Fu, Xu, Ye, Zhang, Xie, Cheng, Zhang, Yang, Xu, and Lin]{qwen25vl}
Shuai Bai, Keqin Chen, Xuejing Liu, Jialin Wang, Wenbin Ge, Sibo Song, Kai Dang, Peng Wang, Shijie Wang, Jun Tang, Humen Zhong, Yuanzhi Zhu, Mingkun Yang, Zhaohai Li, Jianqiang Wan, Pengfei Wang, Wei Ding, Zheren Fu, Yiheng Xu, Jiabo Ye, Xi~Zhang, Tianbao Xie, Zesen Cheng, Hang Zhang, Zhibo Yang, Haiyang Xu, and Junyang Lin.
\newblock Qwen2.5-vl technical report.
\newblock \emph{arXiv preprint arXiv:2502.13923}, 2025.

\bibitem[Bigverdi et~al.(2025)Bigverdi, Luo, Hsieh, Shen, Chen, Shapiro, and Krishna]{bigverdi2025perception}
Mahtab Bigverdi, Zelun Luo, Cheng-Yu Hsieh, Ethan Shen, Dongping Chen, Linda~G Shapiro, and Ranjay Krishna.
\newblock Perception tokens enhance visual reasoning in multimodal language models.
\newblock In \emph{CVPR}, 2025.

\bibitem[Cai et~al.(2024)Cai, Liu, Mustikovela, Meyer, Chai, Park, and Lee]{vipllava}
Mu~Cai, Haotian Liu, Siva~Karthik Mustikovela, Gregory~P Meyer, Yuning Chai, Dennis Park, and Yong~Jae Lee.
\newblock Vip-llava: Making large multimodal models understand arbitrary visual prompts.
\newblock In \emph{CVPR}, 2024.

\bibitem[Chen et~al.(2024{\natexlab{a}})Chen, Li, Sun, Wang, and Chen]{chen2024sam4mllm}
Yi-Chia Chen, Wei-Hua Li, Cheng Sun, Yu-Chiang~Frank Wang, and Chu-Song Chen.
\newblock Sam4mllm: Enhance multi-modal large language model for referring expression segmentation.
\newblock In \emph{ECCV}, 2024{\natexlab{a}}.

\bibitem[Chen et~al.(2024{\natexlab{b}})Chen, Wang, Cao, Liu, Gao, Cui, Zhu, Ye, Tian, Liu, et~al.]{internvl25}
Zhe Chen, Weiyun Wang, Yue Cao, Yangzhou Liu, Zhangwei Gao, Erfei Cui, Jinguo Zhu, Shenglong Ye, Hao Tian, Zhaoyang Liu, et~al.
\newblock Expanding performance boundaries of open-source multimodal models with model, data, and test-time scaling.
\newblock \emph{arXiv preprint arXiv:2412.05271}, 2024{\natexlab{b}}.

\bibitem[Chen et~al.(2024{\natexlab{c}})Chen, Wu, Wang, Su, Chen, Xing, Zhong, Zhang, Zhu, Lu, et~al.]{internvl}
Zhe Chen, Jiannan Wu, Wenhai Wang, Weijie Su, Guo Chen, Sen Xing, Muyan Zhong, Qinglong Zhang, Xizhou Zhu, Lewei Lu, et~al.
\newblock Internvl: Scaling up vision foundation models and aligning for generic visual-linguistic tasks.
\newblock In \emph{CVPR}, 2024{\natexlab{c}}.

\bibitem[Chen et~al.(2020)Chen, Wang, Ma, Wong, and Wu]{cops_ref}
Zhenfang Chen, Peng Wang, Lin Ma, Kwan-Yee~K Wong, and Qi~Wu.
\newblock Cops-ref: A new dataset and task on compositional referring expression comprehension.
\newblock In \emph{CVPR}, 2020.

\bibitem[Cheng et~al.(2022)Cheng, Misra, Schwing, Kirillov, and Girdhar]{mask2former}
Bowen Cheng, Ishan Misra, Alexander~G. Schwing, Alexander Kirillov, and Rohit Girdhar.
\newblock Masked-attention mask transformer for universal image segmentation.
\newblock In \emph{CVPR}, 2022.

\bibitem[Cho et~al.(2025)Cho, Madotto, Mavroudi, Afouras, Nagarajan, Maaz, Song, Ma, Hu, Jain, et~al.]{plm}
Jang~Hyun Cho, Andrea Madotto, Effrosyni Mavroudi, Triantafyllos Afouras, Tushar Nagarajan, Muhammad Maaz, Yale Song, Tengyu Ma, Shuming Hu, Suyog Jain, et~al.
\newblock Perceptionlm: Open-access data and models for detailed visual understanding.
\newblock \emph{arXiv preprint arXiv:2504.13180}, 2025.

\bibitem[Contributors(2023)]{xtuner}
XTuner Contributors.
\newblock Xtuner: A toolkit for efficiently fine-tuning llm.
\newblock \url{https://github.com/InternLM/xtuner}, 2023.

\bibitem[Cordts et~al.(2016)Cordts, Omran, Ramos, Rehfeld, Enzweiler, Benenson, Franke, Roth, and Schiele]{cityscapes}
Marius Cordts, Mohamed Omran, Sebastian Ramos, Timo Rehfeld, Markus Enzweiler, Rodrigo Benenson, Uwe Franke, Stefan Roth, and Bernt Schiele.
\newblock The cityscapes dataset for semantic urban scene understanding.
\newblock In \emph{CVPR}, 2016.

\bibitem[Deng et~al.(2024)Deng, Yu, Wang, Shen, and Chen]{coconut}
Xueqing Deng, Qihang Yu, Peng Wang, Xiaohui Shen, and Liang-Chieh Chen.
\newblock Coconut: Modernizing coco segmentation.
\newblock In \emph{CVPR}, 2024.

\bibitem[Ding et~al.(2023)Ding, Liu, He, Jiang, and Loy]{mevis}
Henghui Ding, Chang Liu, Shuting He, Xudong Jiang, and Chen~Change Loy.
\newblock Mevis: A large-scale benchmark for video segmentation with motion expressions.
\newblock In \emph{ICCV}, 2023.

\bibitem[Guo et~al.(2025)Guo, Wu, Zhu, Leng, Shi, Chen, Fan, Wang, Jiang, Wang, et~al.]{guo2025seed1}
Dong Guo, Faming Wu, Feida Zhu, Fuxing Leng, Guang Shi, Haobin Chen, Haoqi Fan, Jian Wang, Jianyu Jiang, Jiawei Wang, et~al.
\newblock Seed1.5-vl technical report.
\newblock \emph{arXiv preprint arXiv:2505.07062}, 2025.

\bibitem[Guo et~al.(2024)Guo, De~Mello, Yin, Byeon, Cheung, Yu, Luo, and Liu]{guo2024regiongpt}
Qiushan Guo, Shalini De~Mello, Hongxu Yin, Wonmin Byeon, Ka~Chun Cheung, Yizhou Yu, Ping Luo, and Sifei Liu.
\newblock Regiongpt: Towards region understanding vision language model.
\newblock In \emph{CVPR}, pages 13796--13806, 2024.

\bibitem[Hong et~al.(2024)Hong, Wang, Ding, Yu, Lv, Wang, Cheng, Huang, Ji, Xue, et~al.]{cogvlm2}
Wenyi Hong, Weihan Wang, Ming Ding, Wenmeng Yu, Qingsong Lv, Yan Wang, Yean Cheng, Shiyu Huang, Junhui Ji, Zhao Xue, et~al.
\newblock Cogvlm2: Visual language models for image and video understanding.
\newblock \emph{arXiv preprint arXiv:2408.16500}, 2024.

\bibitem[Hu et~al.(2025)Hu, Zhu, Zhang, Cheng, Liu, Liu, Ran, Chen, Liu, and Wang]{groundingsuite}
Rui Hu, Lianghui Zhu, Yuxuan Zhang, Tianheng Cheng, Lei Liu, Heng Liu, Longjin Ran, Xiaoxin Chen, Wenyu Liu, and Xinggang Wang.
\newblock Groundingsuite: Measuring complex multi-granular pixel grounding.
\newblock \emph{arXiv preprint arXiv:2503.10596}, 2025.

\bibitem[Huang et~al.(2025)Huang, Jia, Zhai, Cao, Ye, Zhao, Xu, Hu, and Lin]{visionr1}
Wenxuan Huang, Bohan Jia, Zijie Zhai, Shaosheng Cao, Zheyu Ye, Fei Zhao, Zhe Xu, Yao Hu, and Shaohui Lin.
\newblock Vision-r1: Incentivizing reasoning capability in multimodal large language models.
\newblock \emph{arXiv preprint arXiv:2503.06749}, 2025.

\bibitem[Hurst et~al.(2024)Hurst, Lerer, Goucher, Perelman, Ramesh, Clark, Ostrow, Welihinda, Hayes, Radford, et~al.]{gpt4o}
Aaron Hurst, Adam Lerer, Adam~P Goucher, Adam Perelman, Aditya Ramesh, Aidan Clark, AJ~Ostrow, Akila Welihinda, Alan Hayes, Alec Radford, et~al.
\newblock Gpt-4o system card.
\newblock \emph{arXiv preprint arXiv:2410.21276}, 2024.

\bibitem[Jaech et~al.(2024)Jaech, Kalai, Lerer, Richardson, El-Kishky, Low, Helyar, Madry, Beutel, Carney, et~al.]{o1}
Aaron Jaech, Adam Kalai, Adam Lerer, Adam Richardson, Ahmed El-Kishky, Aiden Low, Alec Helyar, Aleksander Madry, Alex Beutel, Alex Carney, et~al.
\newblock Openai o1 system card.
\newblock \emph{arXiv preprint arXiv:2412.16720}, 2024.

\bibitem[Kazemzadeh et~al.(2014)Kazemzadeh, Ordonez, Matten, and Berg]{refclef}
Sahar Kazemzadeh, Vicente Ordonez, Mark Matten, and Tamara Berg.
\newblock Referitgame: Referring to objects in photographs of natural scenes.
\newblock In \emph{EMNLP}, 2014.

\bibitem[Kirillov et~al.(2023)Kirillov, Mintun, Ravi, Mao, Rolland, Gustafson, Xiao, Whitehead, Berg, Lo, et~al.]{sam}
Alexander Kirillov, Eric Mintun, Nikhila Ravi, Hanzi Mao, Chloe Rolland, Laura Gustafson, Tete Xiao, Spencer Whitehead, Alexander~C Berg, Wan-Yen Lo, et~al.
\newblock Segment anything.
\newblock In \emph{ICCV}, 2023.

\bibitem[Lai et~al.(2024)Lai, Tian, Chen, Li, Yuan, Liu, and Jia]{lisa}
Xin Lai, Zhuotao Tian, Yukang Chen, Yanwei Li, Yuhui Yuan, Shu Liu, and Jiaya Jia.
\newblock Lisa: Reasoning segmentation via large language model.
\newblock In \emph{CVPR}, 2024.

\bibitem[Lan et~al.(2025)Lan, Chen, Zhou, Xu, Ke, Wang, Feng, and Zhang]{lan2024text4seg}
Mengcheng Lan, Chaofeng Chen, Yue Zhou, Jiaxing Xu, Yiping Ke, Xinjiang Wang, Litong Feng, and Wayne Zhang.
\newblock Text4seg: Reimagining image segmentation as text generation.
\newblock \emph{ICLR}, 2025.

\bibitem[Lee et~al.(2022)Lee, Kim, Kim, Cho, and Han]{rq}
Doyup Lee, Chiheon Kim, Saehoon Kim, Minsu Cho, and Wook-Shin Han.
\newblock Autoregressive image generation using residual quantization.
\newblock In \emph{CVPR}, 2022.

\bibitem[Li et~al.(2025{\natexlab{a}})Li, Zhang, Teng, Zhang, Liu, and Lan]{refsam}
Yonglin Li, Jing Zhang, Xiao Teng, Haoyu Zhang, Xinwang Liu, and Long Lan.
\newblock Refsam: Efficiently adapting segmenting anything model for referring video object segmentation.
\newblock \emph{Neural Networks}, 2025{\natexlab{a}}.

\bibitem[Li et~al.(2025{\natexlab{b}})Li, Yu, Huang, Liu, Liang, Liu, Che, Yu, Boyd-Graber, Mi, et~al.]{visionsr1}
Zongxia Li, Wenhao Yu, Chengsong Huang, Rui Liu, Zhenwen Liang, Fuxiao Liu, Jingxi Che, Dian Yu, Jordan Boyd-Graber, Haitao Mi, et~al.
\newblock Self-rewarding vision-language model via reasoning decomposition.
\newblock \emph{arXiv preprint arXiv:2508.19652}, 2025{\natexlab{b}}.

\bibitem[Lian et~al.(2025)Lian, Ding, Ge, Liu, Mao, Li, Pavone, Liu, Darrell, Yala, et~al.]{dam}
Long Lian, Yifan Ding, Yunhao Ge, Sifei Liu, Hanzi Mao, Boyi Li, Marco Pavone, Ming-Yu Liu, Trevor Darrell, Adam Yala, et~al.
\newblock Describe anything: Detailed localized image and video captioning.
\newblock \emph{arXiv preprint arXiv:2504.16072}, 2025.

\bibitem[Lin et~al.(2024)Lin, Wei, An, Gao, Zou, Luo, Huang, Zhang, and Li]{mdpv}
Weifeng Lin, Xinyu Wei, Ruichuan An, Peng Gao, Bocheng Zou, Yulin Luo, Siyuan Huang, Shanghang Zhang, and Hongsheng Li.
\newblock Draw-and-understand: Leveraging visual prompts to enable mllms to comprehend what you want.
\newblock \emph{arXiv preprint arXiv:2403.20271}, 2024.

\bibitem[Lin et~al.(2025)Lin, Wei, An, Ren, Chen, Zhang, Guo, Zhang, Zhang, and Li]{lin2025PAM}
Weifeng Lin, Xinyu Wei, Ruichuan An, Tianhe Ren, Tingwei Chen, Renrui Zhang, Ziyu Guo, Wentao Zhang, Lei Zhang, and Hongsheng Li.
\newblock Perceive anything: Recognize, explain, caption, and segment anything in images and videos.
\newblock \emph{arXiv preprint arXiv:2506.05302}, 2025.

\bibitem[Lin et~al.(2020)Lin, Ding, Zeng, and Tao]{gpsnet}
Xin Lin, Changxing Ding, Jinquan Zeng, and Dacheng Tao.
\newblock Gps-net: Graph property sensing network for scene graph generation.
\newblock In \emph{CVPR}, 2020.

\bibitem[Liu et~al.(2023)Liu, Ding, and Jiang]{gres}
Chang Liu, Henghui Ding, and Xudong Jiang.
\newblock Gres: Generalized referring expression segmentation.
\newblock In \emph{CVPR}, 2023.

\bibitem[Liu et~al.(2025{\natexlab{a}})Liu, Ma, Pu, Qi, Wu, Shan, and Chen]{liu2025unipixel}
Ye~Liu, Zongyang Ma, Junfu Pu, Zhongang Qi, Yang Wu, Ying Shan, and Chang~Wen Chen.
\newblock Unipixel: Unified object referring and segmentation for pixel-level visual reasoning.
\newblock \emph{arXiv preprint arXiv:2509.18094}, 2025{\natexlab{a}}.

\bibitem[Liu et~al.(2025{\natexlab{b}})Liu, Peng, Zhong, Yue, Lu, Yu, and Jia]{segzero}
Yuqi Liu, Bohao Peng, Zhisheng Zhong, Zihao Yue, Fanbin Lu, Bei Yu, and Jiaya Jia.
\newblock Seg-zero: Reasoning-chain guided segmentation via cognitive reinforcement.
\newblock \emph{arXiv preprint arXiv:2503.06520}, 2025{\natexlab{b}}.

\bibitem[Liu et~al.(2025{\natexlab{c}})Liu, Qu, Zhong, Peng, Liu, Yu, and Jia]{visionreasoner}
Yuqi Liu, Tianyuan Qu, Zhisheng Zhong, Bohao Peng, Shu Liu, Bei Yu, and Jiaya Jia.
\newblock Visionreasoner: Unified visual perception and reasoning via reinforcement learning.
\newblock \emph{arXiv preprint arXiv:2505.12081}, 2025{\natexlab{c}}.

\bibitem[Liu et~al.(2025{\natexlab{d}})Liu, Sun, Zang, Dong, Cao, Duan, Lin, and Wang]{visualrft}
Ziyu Liu, Zeyi Sun, Yuhang Zang, Xiaoyi Dong, Yuhang Cao, Haodong Duan, Dahua Lin, and Jiaqi Wang.
\newblock Visual-rft: Visual reinforcement fine-tuning.
\newblock In \emph{ICCV}, 2025{\natexlab{d}}.

\bibitem[Loshchilov and Hutter(2016)]{loshchilov2016sgdr}
Ilya Loshchilov and Frank Hutter.
\newblock Sgdr: Stochastic gradient descent with warm restarts.
\newblock \emph{arXiv preprint arXiv:1608.03983}, 2016.

\bibitem[Loshchilov and Hutter(2017)]{AdamW}
Ilya Loshchilov and Frank Hutter.
\newblock Decoupled weight decay regularization.
\newblock \emph{arXiv preprint arXiv:1711.05101}, 2017.

\bibitem[Ma et~al.(2025)Ma, Chern, Shen, Zhong, and Liu]{ma2025rethinking}
Yan Ma, Steffi Chern, Xuyang Shen, Yiran Zhong, and Pengfei Liu.
\newblock Rethinking rl scaling for vision language models: A transparent, from-scratch framework and comprehensive evaluation scheme.
\newblock \emph{arXiv preprint arXiv:2504.02587}, 2025.

\bibitem[Mao et~al.(2016)Mao, Huang, Toshev, Camburu, Yuille, and Murphy]{refcoco}
Junhua Mao, Jonathan Huang, Alexander Toshev, Oana Camburu, Alan~L Yuille, and Kevin Murphy.
\newblock Generation and comprehension of unambiguous object descriptions.
\newblock In \emph{CVPR}, 2016.

\bibitem[Meng et~al.(2025)Meng, Li, Wang, Tan, Zhang, Kong, Tong, Wang, Teng, Wang, et~al.]{meng2025openo3video}
Jiahao Meng, Xiangtai Li, Haochen Wang, Yue Tan, Tao Zhang, Lingdong Kong, Yunhai Tong, Anran Wang, Zhiyang Teng, Yujing Wang, et~al.
\newblock Open-o3 video: Grounded video reasoning with explicit spatio-temporal evidence.
\newblock \emph{arXiv preprint arXiv:2510.20579}, 2025.

\bibitem[Mentzer et~al.(2023)Mentzer, Minnen, Agustsson, and Tschannen]{fsq}
Fabian Mentzer, David Minnen, Eirikur Agustsson, and Michael Tschannen.
\newblock Finite scalar quantization: Vq-vae made simple.
\newblock \emph{arXiv preprint arXiv:2309.15505}, 2023.

\bibitem[Ouyang et~al.(2022)Ouyang, Wu, Jiang, Almeida, Wainwright, Mishkin, Zhang, Agarwal, Slama, Ray, et~al.]{rlhf}
Long Ouyang, Jeffrey Wu, Xu~Jiang, Diogo Almeida, Carroll Wainwright, Pamela Mishkin, Chong Zhang, Sandhini Agarwal, Katarina Slama, Alex Ray, et~al.
\newblock Training language models to follow instructions with human feedback.
\newblock In \emph{NeurIPS}, 2022.

\bibitem[Qi et~al.(2022)Qi, Kuen, Guo, Shen, Gu, Jia, Lin, and Yang]{entityseg}
Lu~Qi, Jason Kuen, Weidong Guo, Tiancheng Shen, Jiuxiang Gu, Jiaya Jia, Zhe Lin, and Ming-Hsuan Yang.
\newblock High-quality entity segmentation.
\newblock \emph{arXiv preprint arXiv:2211.05776}, 2022.

\bibitem[Rafailov et~al.(2023)Rafailov, Sharma, Mitchell, Manning, Ermon, and Finn]{dpo}
Rafael Rafailov, Archit Sharma, Eric Mitchell, Christopher~D Manning, Stefano Ermon, and Chelsea Finn.
\newblock Direct preference optimization: Your language model is secretly a reward model.
\newblock In \emph{NeurIPS}, 2023.

\bibitem[Rasheed et~al.(2024)Rasheed, Maaz, Shaji, Shaker, Khan, Cholakkal, Anwer, Xing, Yang, and Khan]{glamm}
Hanoona Rasheed, Muhammad Maaz, Sahal Shaji, Abdelrahman Shaker, Salman Khan, Hisham Cholakkal, Rao~M Anwer, Eric Xing, Ming-Hsuan Yang, and Fahad~S Khan.
\newblock Glamm: Pixel grounding large multimodal model.
\newblock In \emph{CVPR}, 2024.

\bibitem[Ravi et~al.(2024)Ravi, Gabeur, Hu, Hu, Ryali, Ma, Khedr, R{\"a}dle, Rolland, Gustafson, et~al.]{sam2}
Nikhila Ravi, Valentin Gabeur, Yuan-Ting Hu, Ronghang Hu, Chaitanya Ryali, Tengyu Ma, Haitham Khedr, Roman R{\"a}dle, Chloe Rolland, Laura Gustafson, et~al.
\newblock Sam 2: Segment anything in images and videos.
\newblock \emph{arXiv preprint arXiv:2408.00714}, 2024.

\bibitem[Schulman et~al.(2017)Schulman, Wolski, Dhariwal, Radford, and Klimov]{ppo}
John Schulman, Filip Wolski, Prafulla Dhariwal, Alec Radford, and Oleg Klimov.
\newblock Proximal policy optimization algorithms.
\newblock \emph{arXiv preprint arXiv:1707.06347}, 2017.

\bibitem[Shao et~al.(2024)Shao, Wang, Zhu, Xu, Song, Bi, Zhang, Zhang, Li, Wu, et~al.]{grpo}
Zhihong Shao, Peiyi Wang, Qihao Zhu, Runxin Xu, Junxiao Song, Xiao Bi, Haowei Zhang, Mingchuan Zhang, YK~Li, Yang Wu, et~al.
\newblock Deepseekmath: Pushing the limits of mathematical reasoning in open language models.
\newblock \emph{arXiv preprint arXiv:2402.03300}, 2024.

\bibitem[Shen et~al.(2025)Shen, Liu, Li, Fang, Ma, Liao, Shen, Zhang, Zhao, Zhang, et~al.]{vlmr1}
Haozhan Shen, Peng Liu, Jingcheng Li, Chunxin Fang, Yibo Ma, Jiajia Liao, Qiaoli Shen, Zilun Zhang, Kangjia Zhao, Qianqian Zhang, et~al.
\newblock Vlm-r1: A stable and generalizable r1-style large vision-language model.
\newblock \emph{arXiv preprint arXiv:2504.07615}, 2025.

\bibitem[Su et~al.(2025)Su, Zhang, Li, Liu, Liao, Pan, Liu, Xing, Sun, Li, et~al.]{padt}
Yongyi Su, Haojie Zhang, Shijie Li, Nanqing Liu, Jingyi Liao, Junyi Pan, Yuan Liu, Xiaofen Xing, Chong Sun, Chen Li, et~al.
\newblock Patch-as-decodable-token: Towards unified multi-modal vision tasks in mllms.
\newblock \emph{arXiv preprint arXiv:2510.01954}, 2025.

\bibitem[Tanaka et~al.(2019)Tanaka, Itamochi, Narioka, Sato, Ushiku, and Harada]{refgta}
Mikihiro Tanaka, Takayuki Itamochi, Kenichi Narioka, Ikuro Sato, Yoshitaka Ushiku, and Tatsuya Harada.
\newblock Generating easy-to-understand referring expressions for target identifications.
\newblock In \emph{ICCV}, 2019.

\bibitem[Tang et~al.(2019)Tang, Zhang, Wu, Luo, and Liu]{vctree}
Kaihua Tang, Hanwang Zhang, Baoyuan Wu, Wenhan Luo, and Wei Liu.
\newblock Learning to compose dynamic tree structures for visual contexts.
\newblock In \emph{CVPR}, 2019.

\bibitem[Team et~al.(2023)Team, Anil, Borgeaud, Alayrac, Yu, Soricut, Schalkwyk, Dai, Hauth, Millican, et~al.]{gemini}
Gemini Team, Rohan Anil, Sebastian Borgeaud, Jean-Baptiste Alayrac, Jiahui Yu, Radu Soricut, Johan Schalkwyk, Andrew~M Dai, Anja Hauth, Katie Millican, et~al.
\newblock Gemini: a family of highly capable multimodal models.
\newblock \emph{arXiv preprint arXiv:2312.11805}, 2023.

\bibitem[Team et~al.(2024)Team, Hong, Yu, Gu, Wang, Gan, Tang, Cheng, Qi, Ji, Pan, Duan, Wang, Wang, Cheng, He, Su, Yang, Pan, Zeng, Wang, Chen, Shi, Pang, Zhang, Yin, Yang, Chen, Xu, Zhu, Chen, Chen, Chen, Lin, Wang, Chen, Lei, Gong, Pan, Liu, Xu, Zhang, Zheng, Yang, Zhong, Huang, Zhao, Xue, Tu, Meng, Zhang, Luo, Hao, Tong, Li, Jia, Liu, Zhang, Lyu, Fan, Huang, Wang, Xue, Wang, Wang, An, Du, Shi, Huang, Niu, Wang, Yue, Li, Zhang, Wang, Wang, Zhang, Xue, Hou, Du, Wang, Zhang, Liu, Xu, Li, Huang, Dong, and Tang]{glm45v}
V~Team, Wenyi Hong, Wenmeng Yu, Xiaotao Gu, Guo Wang, Guobing Gan, Haomiao Tang, Jiale Cheng, Ji~Qi, Junhui Ji, Lihang Pan, Shuaiqi Duan, Weihan Wang, Yan Wang, Yean Cheng, Zehai He, Zhe Su, Zhen Yang, Ziyang Pan, Aohan Zeng, Baoxu Wang, Bin Chen, Boyan Shi, Changyu Pang, Chenhui Zhang, Da~Yin, Fan Yang, Guoqing Chen, Jiazheng Xu, Jiale Zhu, Jiali Chen, Jing Chen, Jinhao Chen, Jinghao Lin, Jinjiang Wang, Junjie Chen, Leqi Lei, Letian Gong, Leyi Pan, Mingdao Liu, Mingde Xu, Mingzhi Zhang, Qinkai Zheng, Sheng Yang, Shi Zhong, Shiyu Huang, Shuyuan Zhao, Siyan Xue, Shangqin Tu, Shengbiao Meng, Tianshu Zhang, Tianwei Luo, Tianxiang Hao, Tianyu Tong, Wenkai Li, Wei Jia, Xiao Liu, Xiaohan Zhang, Xin Lyu, Xinyue Fan, Xuancheng Huang, Yanling Wang, Yadong Xue, Yanfeng Wang, Yanzi Wang, Yifan An, Yifan Du, Yiming Shi, Yiheng Huang, Yilin Niu, Yuan Wang, Yuanchang Yue, Yuchen Li, Yutao Zhang, Yuting Wang, Yu~Wang, Yuxuan Zhang, Zhao Xue, Zhenyu Hou, Zhengxiao Du, Zihan Wang, Peng Zhang, Debing Liu, Bin Xu, Juanzi Li,
  Minlie Huang, Yuxiao Dong, and Jie Tang.
\newblock Glm-4.5v and glm-4.1v-thinking: Towards versatile multimodal reasoning with scalable reinforcement learning.
\newblock \emph{arXiv preprint arXiv:2507.01006}, 2024.

\bibitem[Van Den~Oord et~al.(2017)Van Den~Oord, Vinyals, et~al.]{vqvae}
Aaron Van Den~Oord, Oriol Vinyals, et~al.
\newblock Neural discrete representation learning.
\newblock In \emph{NeurIPS}, 2017.

\bibitem[Wang et~al.(2024{\natexlab{a}})Wang, Ye, Wang, Nie, and Huang]{wang2024Elysium}
Han Wang, Yongjie Ye, Yanjie Wang, Yuxiang Nie, and Can Huang.
\newblock Elysium: Exploring object-level perception in videos via mllm.
\newblock \emph{arXiv preprint arXiv:2403.16558}, 2024{\natexlab{a}}.

\bibitem[Wang et~al.(2025{\natexlab{a}})Wang, Qiao, Jie, Huang, Feng, Zheng, Ma, Lan, and Liang]{xsam}
Hao Wang, Limeng Qiao, Zequn Jie, Zhijian Huang, Chengjian Feng, Qingfang Zheng, Lin Ma, Xiangyuan Lan, and Xiaodan Liang.
\newblock X-sam: From segment anything to any segmentation.
\newblock \emph{arXiv preprint arXiv:2508.04655}, 2025{\natexlab{a}}.

\bibitem[Wang et~al.(2025{\natexlab{b}})Wang, Li, Huang, Wang, Wang, Zhang, Zheng, Bai, Kang, Feng, et~al.]{wang2025traceable}
Haochen Wang, Xiangtai Li, Zilong Huang, Anran Wang, Jiacong Wang, Tao Zhang, Jiani Zheng, Sule Bai, Zijian Kang, Jiashi Feng, et~al.
\newblock Traceable evidence enhanced visual grounded reasoning: Evaluation and methodology.
\newblock \emph{arXiv preprint arXiv:2507.07999}, 2025{\natexlab{b}}.

\bibitem[Wang et~al.(2025{\natexlab{c}})Wang, Wang, Zhang, Zhou, Li, Wang, Tian, Meng, Huang, Mai, et~al.]{gar}
Haochen Wang, Yuhao Wang, Tao Zhang, Yikang Zhou, Yanwei Li, Jiacong Wang, Ye~Tian, Jiahao Meng, Zilong Huang, Guangcan Mai, et~al.
\newblock Grasp any region: Towards precise, contextual pixel understanding for multimodal llms.
\newblock \emph{arXiv preprint arXiv:2510.18876}, 2025{\natexlab{c}}.

\bibitem[Wang et~al.(2025{\natexlab{d}})Wang, Kang, Wang, Jiang, Li, Wu, Wang, Ran, Liang, Feng, et~al.]{wang2025vgr}
Jiacong Wang, Zijian Kang, Haochen Wang, Haiyong Jiang, Jiawen Li, Bohong Wu, Ya~Wang, Jiao Ran, Xiao Liang, Chao Feng, et~al.
\newblock Vgr: Visual grounded reasoning.
\newblock \emph{arXiv preprint arXiv:2506.11991}, 2025{\natexlab{d}}.

\bibitem[Wang et~al.(2023{\natexlab{a}})Wang, Zhang, Chu, Cao, Zhou, Wu, Wang, He, and Lin]{v3det}
Jiaqi Wang, Pan Zhang, Tao Chu, Yuhang Cao, Yujie Zhou, Tong Wu, Bin Wang, Conghui He, and Dahua Lin.
\newblock V3det: Vast vocabulary visual detection dataset.
\newblock In \emph{ICCV}, 2023{\natexlab{a}}.

\bibitem[Wang et~al.(2025{\natexlab{e}})Wang, Wu, Huang, Zheng, and Wang]{wang2025MLLMSeg}
Jingchao Wang, Zhijian Wu, Dingjiang Huang, Yefeng Zheng, and Hong Wang.
\newblock Unlocking the potential of mllms in referring expression segmentation via a light-weight mask decoder.
\newblock \emph{arXiv preprint arXiv:2508.04107}, 2025{\natexlab{e}}.

\bibitem[Wang et~al.(2025{\natexlab{f}})Wang, Lin, Chen, Wang, Cheng, Zhong, Zheng, and Zhao]{alto}
Lingfeng Wang, Hualing Lin, Senda Chen, Tao Wang, Changxu Cheng, Yangyang Zhong, Dong Zheng, and Wuyue Zhao.
\newblock Alto: Adaptive-length tokenizer for autoregressive mask generation.
\newblock \emph{arXiv preprint arXiv:2505.16495}, 2025{\natexlab{f}}.

\bibitem[Wang et~al.(2024{\natexlab{b}})Wang, Bai, Tan, Wang, Fan, Bai, Chen, Liu, Wang, Ge, Fan, Dang, Du, Ren, Men, Liu, Zhou, Zhou, and Lin]{qwen2vl}
Peng Wang, Shuai Bai, Sinan Tan, Shijie Wang, Zhihao Fan, Jinze Bai, Keqin Chen, Xuejing Liu, Jialin Wang, Wenbin Ge, Yang Fan, Kai Dang, Mengfei Du, Xuancheng Ren, Rui Men, Dayiheng Liu, Chang Zhou, Jingren Zhou, and Junyang Lin.
\newblock Qwen2-vl: Enhancing vision-language model's perception of the world at any resolution.
\newblock \emph{arXiv preprint arXiv:2409.12191}, 2024{\natexlab{b}}.

\bibitem[Wang et~al.(2025{\natexlab{g}})Wang, Cheng, Wang, Chen, and Zhao]{himtok}
Tao Wang, Changxu Cheng, Lingfeng Wang, Senda Chen, and Wuyue Zhao.
\newblock Himtok: Learning hierarchical mask tokens for image segmentation with large multimodal model.
\newblock In \emph{ICCV}, 2025{\natexlab{g}}.

\bibitem[Wang et~al.(2025{\natexlab{h}})Wang, Gao, Gu, Pu, Cui, Wei, Liu, Jing, Ye, Shao, et~al.]{internvl35}
Weiyun Wang, Zhangwei Gao, Lixin Gu, Hengjun Pu, Long Cui, Xingguang Wei, Zhaoyang Liu, Linglin Jing, Shenglong Ye, Jie Shao, et~al.
\newblock Internvl3.5: Advancing open-source multimodal models in versatility, reasoning, and efficiency.
\newblock \emph{arXiv preprint arXiv:2508.18265}, 2025{\natexlab{h}}.

\bibitem[Wang et~al.(2023{\natexlab{b}})Wang, Chen, Chen, Wu, Zhu, Zeng, Luo, Lu, Zhou, Qiao, et~al.]{wang2023visionllm}
Wenhai Wang, Zhe Chen, Xiaokang Chen, Jiannan Wu, Xizhou Zhu, Gang Zeng, Ping Luo, Tong Lu, Jie Zhou, Yu~Qiao, et~al.
\newblock Visionllm: Large language model is also an open-ended decoder for vision-centric tasks.
\newblock \emph{NeurIPS}, 2023{\natexlab{b}}.

\bibitem[Wang et~al.(2025{\natexlab{i}})Wang, Ru, Huang, Ji, Zheng, Chen, and Zhou]{argenseg}
Xiaolong Wang, Lixiang Ru, Ziyuan Huang, Kaixiang Ji, Dandan Zheng, Jingdong Chen, and Jun Zhou.
\newblock Argenseg: Image segmentation with autoregressive image generation model.
\newblock In \emph{NeurIPS}, 2025{\natexlab{i}}.

\bibitem[Wang et~al.(2025{\natexlab{j}})Wang, Zhang, Li, Li, Kallidromitis, Kato, Kozuka, and Darrell]{segllm}
XuDong Wang, Shaolun Zhang, Shufan Li, Kehan Li, Konstantinos Kallidromitis, Yusuke Kato, Kazuki Kozuka, and Trevor Darrell.
\newblock Segllm: Multi-round reasoning segmentation with large language models.
\newblock In \emph{ICLR}, 2025{\natexlab{j}}.

\bibitem[Wei et~al.(2024)Wei, Zhong, Tan, Liu, Zhao, Hu, and Yang]{hyperseg}
Cong Wei, Yujie Zhong, Haoxian Tan, Yong Liu, Zheng Zhao, Jie Hu, and Yujiu Yang.
\newblock Hyperseg: Towards universal visual segmentation with large language model.
\newblock \emph{arXiv preprint arXiv:2411.17606}, 2024.

\bibitem[Wei et~al.(2025)Wei, Zhong, Tan, Zeng, Liu, Wang, and Yang]{wei2025instructseg}
Cong Wei, Yujie Zhong, Haoxian Tan, Yingsen Zeng, Yong Liu, Hongfa Wang, and Yujiu Yang.
\newblock Instructseg: Unifying instructed visual segmentation with multi-modal large language models.
\newblock In \emph{ICCV}, pages 20193--20203, 2025.

\bibitem[Wu et~al.(2024)Wu, Chen, Pan, Liu, Liu, Dai, Gao, Ma, Wu, Wang, et~al.]{deepseekvl2}
Zhiyu Wu, Xiaokang Chen, Zizheng Pan, Xingchao Liu, Wen Liu, Damai Dai, Huazuo Gao, Yiyang Ma, Chengyue Wu, Bingxuan Wang, et~al.
\newblock Deepseek-vl2: Mixture-of-experts vision-language models for advanced multimodal understanding.
\newblock \emph{arXiv preprint arXiv:2412.10302}, 2024.

\bibitem[Xu et~al.(2017)Xu, Zhu, Choy, and Fei-Fei]{imp}
Danfei Xu, Yuke Zhu, Christopher~B Choy, and Li~Fei-Fei.
\newblock Scene graph generation by iterative message passing.
\newblock In \emph{CVPR}, 2017.

\bibitem[Yan et~al.(2024)Yan, Wang, Yan, Jiang, Hu, Kang, Xie, and Gavves]{revos}
Cilin Yan, Haochen Wang, Shilin Yan, Xiaolong Jiang, Yao Hu, Guoliang Kang, Weidi Xie, and Efstratios Gavves.
\newblock Visa: Reasoning video object segmentation via large language models.
\newblock In \emph{ECCV}, 2024.

\bibitem[Yang et~al.(2023{\natexlab{a}})Yang, Zhang, Li, Zou, Li, and Gao]{som}
Jianwei Yang, Hao Zhang, Feng Li, Xueyan Zou, Chunyuan Li, and Jianfeng Gao.
\newblock Set-of-mark prompting unleashes extraordinary visual grounding in gpt-4v.
\newblock \emph{arXiv preprint arXiv:2310.11441}, 2023{\natexlab{a}}.

\bibitem[Yang et~al.(2022)Yang, Ang, Guo, Zhou, Zhang, and Liu]{psg}
Jingkang Yang, Yi~Zhe Ang, Zujin Guo, Kaiyang Zhou, Wayne Zhang, and Ziwei Liu.
\newblock Panoptic scene graph generation.
\newblock In \emph{ECCV}, 2022.

\bibitem[Yang et~al.(2025)Yang, Bi, Shen, Guo, and Ma]{pixelweb}
Qi~Yang, Weichen Bi, Haiyang Shen, Yaoqi Guo, and Yun Ma.
\newblock Pixelweb: The first web gui dataset with pixel-wise labels.
\newblock \emph{arXiv preprint arXiv:2504.16419}, 2025.

\bibitem[Yang et~al.(2023{\natexlab{b}})Yang, Qu, Lai, Tian, Peng, Liu, and Jia]{yang2023lisa++}
Senqiao Yang, Tianyuan Qu, Xin Lai, Zhuotao Tian, Bohao Peng, Shu Liu, and Jiaya Jia.
\newblock Lisa++: An improved baseline for reasoning segmentation with large language model.
\newblock \emph{arXiv preprint arXiv:2312.17240}, 2023{\natexlab{b}}.

\bibitem[You and Wu(2025)]{segr1}
Zuyao You and Zuxuan Wu.
\newblock Seg-r1: Segmentation can be surprisingly simple with reinforcement learning.
\newblock \emph{arXiv preprint arXiv:2506.22624}, 2025.

\bibitem[Young et~al.(2014)Young, Lai, Hodosh, and Hockenmaier]{flickr30k}
Peter Young, Alice Lai, Micah Hodosh, and Julia Hockenmaier.
\newblock From image descriptions to visual denotations: New similarity metrics for semantic inference over event descriptions.
\newblock In \emph{TACL}, 2014.

\bibitem[Yu et~al.(2016)Yu, Poirson, Yang, Berg, and Berg]{refcoco_p_g}
Licheng Yu, Patrick Poirson, Shan Yang, Alexander~C Berg, and Tamara~L Berg.
\newblock Modeling context in referring expressions.
\newblock In \emph{ECCV}, 2016.

\bibitem[Yuan et~al.(2025{\natexlab{a}})Yuan, Li, Zhang, Huang, Xu, Ji, Tong, Qi, Feng, and Yang]{sa2va}
Haobo Yuan, Xiangtai Li, Tao Zhang, Zilong Huang, Shilin Xu, Shunping Ji, Yunhai Tong, Lu~Qi, Jiashi Feng, and Ming-Hsuan Yang.
\newblock Sa2va: Marrying sam2 with llava for dense grounded understanding of images and videos.
\newblock \emph{arXiv preprint arXiv:2501.04001}, 2025{\natexlab{a}}.

\bibitem[Yuan et~al.(2025{\natexlab{b}})Yuan, Sun, Li, Zhang, Deng, Ding, Qi, Wang, Li, and Yang]{yuan2025visual}
Haobo Yuan, Yueyi Sun, Yanwei Li, Tao Zhang, Xueqing Deng, Henghui Ding, Lu~Qi, Anran Wang, Xiangtai Li, and Ming-Hsuan Yang.
\newblock Visual reasoning tracer: Object-level grounded reasoning benchmark.
\newblock \emph{arXiv preprint arXiv:2512.05091}, 2025{\natexlab{b}}.

\bibitem[Yuan et~al.(2024)Yuan, Li, Liu, Tang, Luo, Qin, Zhang, and Zhu]{osprey}
Yuqian Yuan, Wentong Li, Jian Liu, Dongqi Tang, Xinjie Luo, Chi Qin, Lei Zhang, and Jianke Zhu.
\newblock Osprey: Pixel understanding with visual instruction tuning.
\newblock In \emph{CVPR}, 2024.

\bibitem[Yuan et~al.(2025{\natexlab{c}})Yuan, Zhang, Li, Cheng, Zhang, Li, Li, Zhao, Zhang, Zhuang, et~al.]{videorefer}
Yuqian Yuan, Hang Zhang, Wentong Li, Zesen Cheng, Boqiang Zhang, Long Li, Xin Li, Deli Zhao, Wenqiao Zhang, Yueting Zhuang, et~al.
\newblock Videorefer suite: Advancing spatial-temporal object understanding with video llm.
\newblock In \emph{CVPR}, 2025{\natexlab{c}}.

\bibitem[Yuan et~al.(2025{\natexlab{d}})Yuan, Zhang, Li, Wang, Li, Li, Xiao, Zhang, and Ooi]{pixelrefer}
Yuqian Yuan, Wenqiao Zhang, Xin Li, Shihao Wang, Kehan Li, Wentong Li, Jun Xiao, Lei Zhang, and Beng~Chin Ooi.
\newblock Pixelrefer: A unified framework for spatio-temporal object referring with arbitrary granularity.
\newblock \emph{arXiv preprint arXiv:2510.23603}, 2025{\natexlab{d}}.

\bibitem[Zellers et~al.(2018)Zellers, Yatskar, Thomson, and Choi]{motif}
Rowan Zellers, Mark Yatskar, Sam Thomson, and Yejin Choi.
\newblock Neural motifs: Scene graph parsing with global context.
\newblock In \emph{CVPR}, 2018.

\bibitem[Zhang et~al.(2023)Zhang, Xu, Mo, and Kong]{invig}
Hanbo Zhang, Jie Xu, Yuchen Mo, and Tao Kong.
\newblock Invig: Benchmarking interactive visual grounding with 500k human-robot interactions.
\newblock \emph{arXiv preprint arXiv:2310.12147}, 2023.

\bibitem[Zhang et~al.(2024{\natexlab{a}})Zhang, You, Dufter, Zhang, Chen, Chen, Fu, Wang, Chang, Gan, et~al.]{ferret2}
Haotian Zhang, Haoxuan You, Philipp Dufter, Bowen Zhang, Chen Chen, Hong-You Chen, Tsu-Jui Fu, William~Yang Wang, Shih-Fu Chang, Zhe Gan, et~al.
\newblock Ferret-v2: An improved baseline for referring and grounding with large language models.
\newblock \emph{arXiv preprint arXiv:2404.07973}, 2024{\natexlab{a}}.

\bibitem[Zhang et~al.(2024{\natexlab{b}})Zhang, Li, Fei, Yuan, Wu, Ji, Loy, and Yan]{zhang2024omg}
Tao Zhang, Xiangtai Li, Hao Fei, Haobo Yuan, Shengqiong Wu, Shunping Ji, Chen~Change Loy, and Shuicheng Yan.
\newblock Omg-llava: Bridging image-level, object-level, pixel-level reasoning and understanding.
\newblock \emph{NeurIPS}, 2024{\natexlab{b}}.

\bibitem[Zhang et~al.(2025{\natexlab{a}})Zhang, Li, Huang, Li, Lei, Deng, Chen, Ji, and Feng]{zhang2025pixel}
Tao Zhang, Xiangtai Li, Zilong Huang, Yanwei Li, Weixian Lei, Xueqing Deng, Shihao Chen, Shunping Ji, and Jiashi Feng.
\newblock Pixel-sail: Single transformer for pixel-grounded understanding.
\newblock \emph{arXiv preprint arXiv:2504.10465}, 2025{\natexlab{a}}.

\bibitem[Zhang et~al.(2025{\natexlab{b}})Zhang, Wei, Chen, Yu, Luo, and Ji]{vectorllm}
Tao Zhang, Shiqing Wei, Shihao Chen, Wenling Yu, Muying Luo, and Shunping Ji.
\newblock Vectorllm: Human-like extraction of structured building contours vis multimodal llms.
\newblock \emph{arXiv preprint arXiv:2507.04664}, 2025{\natexlab{b}}.

\bibitem[Zhang et~al.(2024{\natexlab{c}})Zhang, Cheng, Zhu, Hu, Liu, Liu, Ran, Chen, Liu, and Wang]{zhang2024evf-sam}
Yuxuan Zhang, Tianheng Cheng, Lianghui Zhu, Rui Hu, Lei Liu, Heng Liu, Longjin Ran, Xiaoxin Chen, Wenyu Liu, and Xinggang Wang.
\newblock Evf-sam: Early vision-language fusion for text-prompted segment anything model.
\newblock \emph{arXiv preprint arXiv:2406.20076}, 2024{\natexlab{c}}.

\bibitem[Zheng et~al.(2025)Zheng, Lu, Wang, Feng, Kuang, and Xiong]{easyr1}
Yaowei Zheng, Junting Lu, Shenzhi Wang, Zhangchi Feng, Dongdong Kuang, and Yuwen Xiong.
\newblock Easyr1: An efficient, scalable, multi-modality rl training framework.
\newblock \url{https://github.com/hiyouga/EasyR1}, 2025.

\bibitem[Zhou et~al.(2017)Zhou, Zhao, Puig, Fidler, Barriuso, and Torralba]{ade20k}
Bolei Zhou, Hang Zhao, Xavier Puig, Sanja Fidler, Adela Barriuso, and Antonio Torralba.
\newblock Scene parsing through ade20k dataset.
\newblock In \emph{CVPR}, 2017.

\bibitem[Zhou et~al.(2025)Zhou, Zhang, Xu, Chen, Zhou, Tong, Ji, Zhang, Qi, and Li]{mmvm}
Yikang Zhou, Tao Zhang, Shilin Xu, Shihao Chen, Qianyu Zhou, Yunhai Tong, Shunping Ji, Jiangning Zhang, Lu~Qi, and Xiangtai Li.
\newblock Are they the same? exploring visual correspondence shortcomings of multimodal llms.
\newblock In \emph{ICCV}, 2025.

\bibitem[Zhu et~al.(2025{\natexlab{a}})Zhu, Wang, Chen, Liu, Ye, Gu, Tian, Duan, Su, Shao, et~al.]{internvl3}
Jinguo Zhu, Weiyun Wang, Zhe Chen, Zhaoyang Liu, Shenglong Ye, Lixin Gu, Hao Tian, Yuchen Duan, Weijie Su, Jie Shao, et~al.
\newblock Internvl3: Exploring advanced training and test-time recipes for open-source multimodal models.
\newblock \emph{arXiv preprint arXiv:2504.10479}, 2025{\natexlab{a}}.

\bibitem[Zhu et~al.(2025{\natexlab{b}})Zhu, Ouyang, Zhang, Cheng, Hu, Shen, Ran, Chen, Yu, Liu, et~al.]{lens}
Lianghui Zhu, Bin Ouyang, Yuxuan Zhang, Tianheng Cheng, Rui Hu, Haocheng Shen, Longjin Ran, Xiaoxin Chen, Li~Yu, Wenyu Liu, et~al.
\newblock Lens: Learning to segment anything with unified reinforced reasoning.
\newblock \emph{arXiv preprint arXiv:2508.14153}, 2025{\natexlab{b}}.

\bibitem[Zhu et~al.(2016)Zhu, Groth, Bernstein, and Fei-Fei]{visual7w}
Yuke Zhu, Oliver Groth, Michael Bernstein, and Li~Fei-Fei.
\newblock Visual7w: Grounded question answering in images.
\newblock In \emph{CVPR}, 2016.

\end{thebibliography}

\end{document}